\documentclass{article}

\usepackage[dblblindworkshop,final]{neurips_2026}

\usepackage[utf8]{inputenc} 
\usepackage[T1]{fontenc}    
\usepackage{xcolor}         
\definecolor{LinkBlue}{HTML}{1F4E79}
\definecolor{CiteGreen}{HTML}{00843D}
\definecolor{UrlBlue}{HTML}{176B87}
  \usepackage[
    colorlinks=true,
    linkcolor=LinkBlue,
    citecolor=CiteGreen,
    urlcolor=UrlBlue
  ]{hyperref}

\usepackage{url}            
\usepackage{booktabs}       
\usepackage{amsfonts}       
\usepackage{nicefrac}       
\usepackage{microtype}      
\usepackage{tikz}           
\usepackage{pgfplots}       
\usepackage{amsmath,amssymb,amsthm,mathtools,bbm}

\pgfplotsset{compat=1.17}
\usepgfplotslibrary{groupplots}
\usepackage{wrapfig}        
\usepackage{placeins} 

\title{Fitting Vision Adapters at Frontier Scales}
\workshoptitle{Grounded and Faithful Vision-Language Models for Real-World Deployment (VLM4RWD)}
\author{%
  \textbf{Jaehoon Lee}\thanks{Equal contribution.} \\
  Base Labs \\
  {\small\texttt{jaehoon.lee@baseten.co}}
  \And
  \textbf{Harry Partridge}\footnotemark[1] \\
  Base Labs \\
  {\small\texttt{harry.partridge@baseten.co}}
  \And
  \textbf{Mudith Jayasekara} \\
  Base Labs \\
  {\small\texttt{mudith@baseten.co}}
  \AND
  \textbf{Charles O'Neill} \\
  Base Labs \\
  {\small\texttt{charles.oneill@baseten.co}}
  \And
  \textbf{Max Kirkby} \\
  Base Labs \\
  {\small\texttt{max.kirkby@baseten.co}}
  \And
  \textbf{Michael Psenka} \\
  Base Labs \\
  {\small\texttt{psenka@baseten.co}}
}

\begin{document}

\maketitle

\begin{abstract}
Training a small projector between a frozen vision encoder and language model is an established approach to multimodal learning. As the parameter count of language models scales dramatically, we revisit which vision capabilities this approach can add while keeping their pretrained weights fixed.
Here we train a 50M parameter projector from the vision encoder of Kimi K2.6 to GLM 5.2 and 5.3, both models without native vision capabilities, and further present a reproducible recipe for training these adapters at scale. We study the following: (a) how vision capabilities of multimodal models scale as purely the language model side scales, and (b) what specific vision capabilities are able to be imbued into a pure language model at scale, and which ones remain limited. We evaluate on MMMU-Pro and BLINK, examining both overall performance and results on individual visual tasks.
\end{abstract}

\section{Introduction}
\label{sec:intro}

Most current vision-language models are built by updating a vision
encoder and a language model together on multimodal data. An even
earlier approach trains only a small projector between a frozen encoder and a
frozen language model. The lower performance of this method relative to joint training has been attributed to the projector's limited
capacity \citep{dai2024nvlm}. Since then, 
the scale of open-source LLMs has grown drastically, so we revisit which visual capabilities a small projector can add while keeping the language model frozen.

Our main empirical finding is that a relatively small trained projector,
50M parameters, is able to bridge a 400M-parameter vision encoder into a
743B-parameter language model, namely GLM-5.2. This is achieved through a
token-alignment initialization, a synthetic SFT pipeline, and an
image-based math RLVR training run, described in detail in
Section~\ref{sec:method}.

From this basis, we study two questions. (a) How do vision capabilities scale as purely the
language side scales? The answer at the frontier is: not much further. Two
unrelated frontier readers land within a point of each other (MMMU-Pro
$56.8$ for GLM-5.2 vs $56.9$ for Nemotron-3-Ultra), which places the
ceiling in the task rather than the reader. (b) At larger language model scale, which specific vision
capabilities can be imbued into a pure language model this way, and which
remain stubborn?

Our training method also produced findings we did not anticipate. Across all tested Qwen3 runs and ablations, training loss plateaus before falling sharply. The MoE runs show no comparable plateau, although their training setups differ (Figure~\ref{fig:moe-training}). Further, SFT with empty reasoning blocks suppresses reasoning output; reinforcement learning on a verifiable visual task restores it across the tested model sizes. 

We present results on MMMU-Pro~\citep{yue2024mmmupro} and BLINK~\citep{fu2024blink} for overall and specific vision capabilities.

    
    

    

\section{Problem formulation}
\label{sec:formulation}

The model has three parts. A pretrained vision encoder $E_{\phi}$ maps an
image $x$ to $n$ patch embeddings $V = E_{\phi}(x) \in \mathbb{R}^{n \times
d_v}$; a small MLP projector $P_{\theta}$ maps each patch embedding
independently into the language model's embedding space, $Z = P_{\theta}(V)
\in \mathbb{R}^{n \times d}$; and a pretrained text-only language model
$f_{\psi}$ consumes the projected patches concatenated with the embedded
question $q$ exactly as if they were token embeddings,
\begin{equation}
  p_{\theta,\phi}(a \mid x, q)
  =
    f_{\psi}\!\left(a \,\middle|\,
      \big[\, Z;\ \mathrm{emb}(q)\,\big]\right),
  \label{eq:model}
\end{equation}
where $\mathrm{emb}(\cdot)$ is the language model's own frozen embedding
matrix and $[\,\cdot\,;\,\cdot\,]$ is sequence concatenation. Within each experiment, the vision encoder and language model parameters
remain fixed, with only the projector parameters $\theta$ trained.
We call $f_{\psi}$ the \emph{reader} (e.g. \textit{the prior}, when the question is what it answers without the image), and $E_{\phi}$ and $P_{\theta}$
together the \emph{perception stack}.

\paragraph{Training.}
During supervised fine-tuning (SFT), we train the projector on
image--question--answer triples
$\mathcal{D} = \{(x_i, q_i, a_i)\}_{i=1}^{N}$ by minimizing the
cross-entropy of the answer tokens,
\begin{equation}
  \mathcal{L}(\theta, \phi)
  = -\frac{1}{N} \sum_{i=1}^{N}
    \log p_{\theta,\phi}(a_i \mid x_i, q_i),
  \label{eq:loss}
\end{equation}
over $\theta$ only. All data, initialization, optimization, and RL are specified in Section~\ref{sec:method}.

\paragraph{Evaluation.}
We evaluate on MMMU \citep{yue2024mmmu}, MMMU-Pro
\citep{yue2024mmmupro}, and all 14 BLINK subtasks
\citep{fu2024blink} (6 single-image, 8 multi-image).
For the GLM-5.2 and Nemotron-3-Ultra results in
Table~\ref{tab:main}, we use generative evaluation with temperature
$1.0$, top-$p$ $0.95$, maximum reasoning effort, a
$131{,}072$-token generation cap, and seed 42.
The official parser extracts answers, with seeded random-choice
fallback for unparseable outputs; fallback rates are reported in
Appendix~\ref{app:training}.
Each vision condition is paired with a \emph{blind} control using
the same model and text prompt without images.
We report accuracy and vision-minus-blind differences to measure
the net effect of providing images under this protocol.
The Qwen3 comparison follows the decoding configuration in
Table~\ref{tab:rung-rl}.

\paragraph{Targeted adapter training.}
For Table~\ref{tab:capability-matrix}, we further train the
GLM-5.2 projector on examples of specific visual tasks, keeping
the encoder and language model frozen. For BLINK evaluation,
we combine each question's images into a single image and score the provided answer choices by their likelihood.

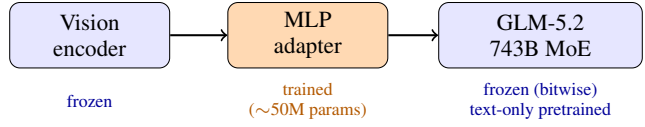
\begin{figure}[t]
  \centering
  \resizebox{0.6\textwidth}{!}{%
  \begin{tikzpicture}[
      box/.style={draw, rounded corners, minimum width=2.2cm, minimum height=0.9cm, align=center, font=\small},
      frozen/.style={box, fill=blue!10},
      trained/.style={box, fill=orange!30},
      lbl/.style={font=\scriptsize, align=center}]
    \node[frozen] (enc) at (0,0) {Vision\\encoder};
    \node[trained] (mlp) at (3.0,0) {MLP\\adapter};
    \node[frozen, minimum width=2.8cm] (llm) at (6.2,0) {GLM-5.2\\743B MoE};
    \draw[->, thick] (enc) -- (mlp);
    \draw[->, thick] (mlp) -- (llm);
    \node[lbl, text=blue!60!black] at (0,-0.9) {frozen};
    \node[lbl, text=orange!70!black] at (3.0,-0.9) {trained\\($\sim$50M params)};
    \node[lbl, text=blue!60!black] at (6.2,-0.9) {frozen (bitwise)\\text-only pretrained};
  \end{tikzpicture}}
  \caption{Primary training setup. Only the 50M parameter MLP adapter is trained; the 743B pure language model remains frozen at its pretraining state.}
  \label{fig:setup}
\end{figure}
\paragraph{Robustness evaluation.}
We further measure how GLM-5.3 accuracy on MMMU-Pro changes when we
add noise or alter image brightness or saturation. We compare each condition with the original images (and same frozen model and generation settings). Appendix~B gives further details. 

\section{Training methodology}
\label{sec:method}
We train projectors for the Qwen3 family (4B, 8B, 14B, and 32B),
GLM-5.2 (743B total parameters), and Nemotron-3-Ultra (550B). Each model is joined to the same frozen vision encoder through an
approximately 50M parameter projector. As described, we train only $P_{\theta}$ and keep the vision encoder and language model frozen. All Qwen3
experiments use the same training configuration (across model sizes). Other experiments with GLM-5.2 and Nemotron-3-Ultra use different configurations, namely in their trainers,
reward compositions, and judges (Figure~\ref{fig:moe-training}). We additionally train a GLM-5.3 adapter using the same GLM-5.2 recipe for
the MMMU-Pro image dependence and robustness study in
Section~\ref{sec:robustness}. Training details for this extension are provided in
Appendix~\ref{app:glm53-robustness}.

\subsection{Data}
\label{sec:method-data}

\paragraph{Supervised fine-tuning.}
We train on $131{,}072$ multi-turn conversations using images from
13 public sources. Kimi-K3 \citep{kimiteam2026kimik3} generates the
question--answer text from each image. We record the source, license,
and image checksum for each example. We shuffle the conversations
once with a fixed random seed and use the same order across runs.

\paragraph{Reinforcement learning.}
The RL task set is $16{,}384$ competition-style math problems (the
Apache-2.0 \texttt{default} split of OpenR1-Math-220k
\citep{openr1math2025}), each \emph{rendered to an image}: the entire problem
statement enters the model through the perception stack, and the text
prompt only asks the model to solve the problem in the image. Each problem's
reference answer is derived by a strong multimodal reasoning model (GPT-5.6 Luna on maximum reasoning effort) solving
the same image independently, without access to the source answer; only problems passing this audit are kept. This gives a
verifiable reward on a task that cannot be solved without reading the
image.

\subsection{Supervised fine-tuning}
\label{sec:method-sft}

\paragraph{Objective.}
SFT minimizes Eq.~\ref{eq:loss} over $\theta$ only. Conversations are
serialized with the reader's chat template, and every assistant turn is
rendered with an \emph{empty reasoning block},
\begin{equation}
  y \;=\; \texttt{<think>}\ \texttt{</think>}\ a\ \texttt{<eot>},
  \label{eq:render}
\end{equation}
so the supervised behavior is to answer immediately. The loss covers the
assistant span only. Section~\ref{sec:results} shows this convention has a
strong side effect: the trained adapter suppresses the reader's reasoning
mode, which the RL stage must undo.

\paragraph{Initialization by embedding alignment.}
Rather than initializing $P_{\theta}$ randomly, we retarget a
\emph{source} projector $P_{\mathrm{src}}$ trained against a different
reader (the encoder's original companion model). Both readers share the
base tokenizer ($|\mathcal{T}| = 95{,}369$ tokens) but embed it in
different spaces, so we fit an affine map $g(v) = Wv + c$ from the source
reader's token embeddings $e^{\mathrm{src}}_t$ to the target reader's
$e^{\mathrm{tgt}}_t$ by ordinary least squares over the shared vocabulary,
\begin{equation}
  W^{*}, c^{*}
  \;=\;
  \arg\min_{W, c}\;
  \frac{\sum_{t \in \mathcal{T}}
        \bigl\| W e^{\mathrm{src}}_t + c - e^{\mathrm{tgt}}_t \bigr\|^2}
       {\sum_{t \in \mathcal{T}}
        \bigl\| e^{\mathrm{tgt}}_t \bigr\|^2},
  \label{eq:align}
\end{equation}
and fold $g$ into the source projector's linear output layer, which gives a
native projector checkpoint for the target reader with no new parameters
(derivation in Appendix~\ref{app:align}); the fit is deterministic and
takes minutes on one GPU. The residual of Eq.~\ref{eq:align} is itself a
finding: across the Kimi$\to$Qwen3 pairs the validation relative MSE is
$0.77$--$0.89$ (cosine similarity $0.36$--$0.50$), so cross-family token
geometries are far from affinely equivalent. The initialization provides a
coarse orientation, and gradient descent does the remainder
(Appendix~\ref{sec:ablations}).

\subsection{Reinforcement learning}
\label{sec:method-rl}

\paragraph{Sampling.}
RL is on-policy: each update draws $64$ problems and samples $G{=}8$
responses per problem from the current policy (temperature $1.0$, top-$p$
$0.95$), with rollouts rendered in the reader's reasoning mode (for the
Qwen3 rungs, the tokenizer's \texttt{/think} switch; recorded in every run
manifest) and capped at $6{,}144$ tokens.

\paragraph{Reward.}
Each response $y$ receives
\begin{equation}
  R(y) \;=\;
  \mathbbm{1}\!\left[\mathrm{valid}(y)\right]
  \Bigl(
    w_f\,\mathbbm{1}[\mathrm{format}]
    + w_r\,\mathbbm{1}[\mathrm{reason}]
    + w_c\,\mathbbm{1}[\mathrm{correct}]
  \Bigr).
  \label{eq:reward}
\end{equation}
We set $(w_f, w_r, w_c) = (0.25,\, 0.25,\, 0.5)$, where $\mathrm{format}$ requires exactly one closed reasoning block,
$\mathrm{reason}$ requires it to be non-empty, $\mathrm{correct}$ requires
the final answer to match the reference (equivalence judged by an LLM
judge - GPT-5.6 Luna) with a natural stop, and $\mathrm{valid}(y)$ gates all positive
terms on non-empty reasoning and non-empty final-answer channels. A
response that overruns the cap typically truncates before its reasoning
block closes and then earns zero, which is the recipe's length pressure.

\paragraph{Objective.}
Advantages are group-normalized,
$\hat{A}_i = (R_i - \mu_G)/(\sigma_G + \epsilon)$, and the policy is
updated with CISPO \citep{minimax2025m1}, a token-level clipped
importance-sampling variant of group-relative policy optimization
\citep{shao2024deepseekmath}; the objective is written out in
Appendix~\ref{app:cispo}. Only $\theta$ is updated; the frozen reader and encoder
are shared between the actor and the rollout engine, and
actor--rollout log-probability agreement is verified before every launch.
Each Qwen3 rung trains for $150$ updates with a checkpoint per update;
the MoE readers use their released RL checkpoints ($154$ and $118$
updates).

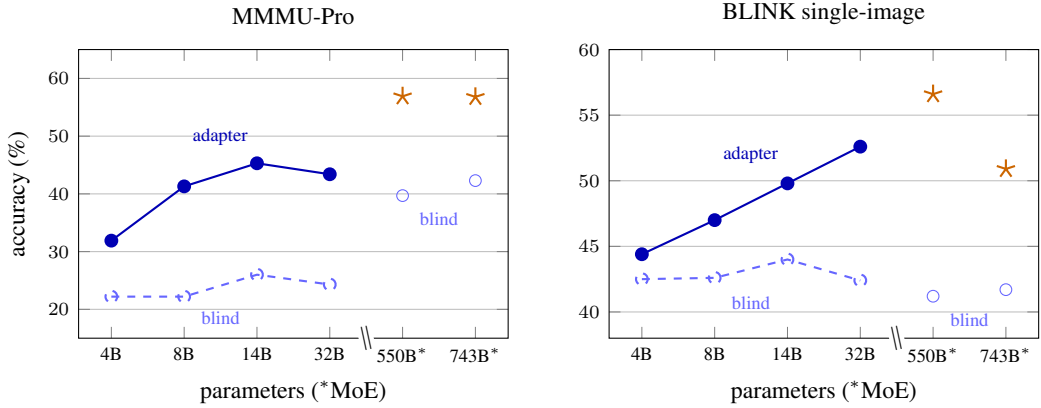
\begin{figure}[t]
  \centering
  \begin{tikzpicture}
  \begin{groupplot}[
      group style={group size=2 by 1, horizontal sep=38pt},
      width=0.52\textwidth, height=5.4cm,
      xmin=0.55, xmax=6.45, ymajorgrids, clip=false,
      xtick={1,2,3,4,5,6},
      xticklabels={4B,8B,14B,32B,550B$^{*}$,743B$^{*}$},
      x tick label style={font=\scriptsize},
      y tick label style={font=\scriptsize},
      label style={font=\small},
      xlabel={parameters ($^{*}$MoE)}]
    \nextgroupplot[title={\small MMMU-Pro}, ylabel={accuracy (\%)}, ymin=15, ymax=65, ytick={20,30,40,50,60}]
      \addplot[blue!70!black, thick, mark=*, mark size=2.2pt]
        coordinates {(1,31.9) (2,41.3) (3,45.3) (4,43.4)};
      \addplot[blue!60, thick, dashed, mark=o, mark size=2.2pt]
        coordinates {(1,22.2) (2,22.2) (3,26.0) (4,24.3)};
      \addplot[only marks, mark=star, mark size=3.6pt, thick, orange!80!black]
        coordinates {(5,56.9) (6,56.8)};
      \addplot[only marks, mark=o, mark size=2.2pt, blue!60]
        coordinates {(5,39.7) (6,42.3)};
      \node[font=\scriptsize, blue!70!black, anchor=south east] at (axis cs:3,46.8) {adapter};
      \node[font=\scriptsize, blue!60, anchor=north] at (axis cs:2.5,21.2) {blind};
      \node[font=\scriptsize, blue!60, anchor=north] at (axis cs:5.5,38.4) {blind};
      \node[font=\small, fill=white, inner sep=0.5pt, rotate=12] at (axis cs:4.5,15) {$\parallel$};
    \nextgroupplot[title={\small BLINK single-image}, ymin=38, ymax=60, ytick={40,45,50,55,60}]
      \addplot[blue!70!black, thick, mark=*, mark size=2.2pt]
        coordinates {(1,44.4) (2,47.0) (3,49.8) (4,52.6)};
      \addplot[blue!60, thick, dashed, mark=o, mark size=2.2pt]
        coordinates {(1,42.5) (2,42.6) (3,44.0) (4,42.4)};
      \addplot[only marks, mark=star, mark size=3.6pt, thick, orange!80!black]
        coordinates {(5,56.6) (6,50.9)};
      \addplot[only marks, mark=o, mark size=2.2pt, blue!60]
        coordinates {(5,41.2) (6,41.7)};
      \node[font=\scriptsize, blue!70!black, anchor=south east] at (axis cs:3,50.6) {adapter};
      \node[font=\scriptsize, blue!60, anchor=north] at (axis cs:2.5,41.9) {blind};
      \node[font=\scriptsize, blue!60, anchor=north] at (axis cs:5.5,40.6) {blind};
      \node[font=\small, fill=white, inner sep=0.5pt, rotate=12] at (axis cs:4.5,38) {$\parallel$};
  \end{groupplot}
  \end{tikzpicture}
  \caption{\textbf{Visual capability for growing language model sizes}, with the same encoder and
  $\approx$50M-parameter adapter method throughout. MMMU-Pro rises steeply
  to 14B and plateaus by 32B, with MoE  about 12 points above
  the dense plateau; blind floors stay flat within each family. BLINK
  single-image saturates at every scale. The break ($\parallel$)
  compresses the axis between 32B and the MoE.}
  \label{fig:scalinglaw}
\end{figure}


\newcommand{\dummynote}{\textbf{\textcolor{red}{[Fabricated placeholder data for layout only.]}}\ }
\newcommand{\fakewatermark}{\node[rotate=18, scale=1.6, text=red!60, opacity=0.5]
  at (rel axis cs:0.5,0.55) {FAKE DATA};}

\section{Results}
\label{sec:results}
We compare performance across language-model sizes (Section~\ref{sec:scaling}) and examine which visual tasks improve
with image input in GLM-5.2 (Section~\ref{sec:blinksplit}). We also
test whether further projector training improves specific GLM-5.2
capabilities (Table~\ref{tab:capability-matrix}). Finally, GLM-5.3 extends
the study with an evaluation of image dependence and robustness
on MMMU-Pro (Section~\ref{sec:robustness}). Interpretability
experiments on GLM-5.2 are presented in Section~\ref{sec:interp}. See Appendix~\ref{app:training} for hyperparameter details.


\subsection{Scaling laws}
\label{sec:scaling}

We train projectors for frozen Qwen3 models at 4B, 8B, 14B, and 32B using the same encoder, projector architecture,
training data, initialization, and hyperparameters at every
scale (Section~\ref{sec:method}). 

We make two observations from this training (for curves see Appendix~\ref{app:training}, Figure~\ref{fig:losslaw}). First, final SFT loss improves with reader scale from 4B to 14B (0.230, 0.178, 0.163); 32B ends slightly higher (0.180), consistent
with its later grokking transition (around step 1{,}100 versus 700 for
8B/14B) leaving fewer post-transition steps, plausibly because more
parameters take longer to grok. Second, loss shows a grokking-like delayed transition ~\citep{power2022grokking}, with each run reaching a plateau before the loss falls sharply (e.g. 14B: from loss
$0.75$ to $0.50$ in 27 steps). 

To investigate this plateau, we vary image-token attention mask and position in a $2\times2$ ablation at 14B. All four configurations show the same plateau followed by a sharp
loss decline. An image-removal control also shows that the
plateau checkpoints already benefit from image input (Appendix~\ref{sec:ablations}).

Table~\ref{tab:rung-rl} reports the Qwen3 benchmark results,
and Figure~\ref{fig:scalinglaw} compares them with the frontier MoE models. The MoE models use different training configurations (Figure~\ref{fig:moe-training}), so these
comparisons do not isolate the effect of model size. Pre-RL results are reported in
Table~\ref{tab:sft-think-baselines}.

\begin{table}[t]
  \caption{\textbf{Benchmark results} Accuracy (\%); chance = random-guess
  rate. Artifacts served and scored with the official generative evaluator
  (decode configuration of Section~\ref{sec:formulation}); blind = same
  served system, no image. 95\% binomial CIs span $\pm 2.3$--$3.4$;
  parse-fallback rates are in Appendix~\ref{app:training}.}
  \label{tab:main}
  \centering
  \small
  \begin{tabular}{lcc cc cc}
    \toprule
    & & & \multicolumn{2}{c}{GLM-5.2} & \multicolumn{2}{c}{Nemotron-3-Ultra} \\
    \cmidrule(lr){4-5}\cmidrule(lr){6-7}
    & $n$ & Chance & Blind & Vision & Blind & Vision \\
    \midrule
    MMMU (standard, val)    & 805  & 26.4 & $63.6$ & $71.4$ & $60.9$ & $67.3$ \\
    MMMU-Pro (standard)     & 1730 & 12.2 & $42.3$ & $56.8$ & $39.7$ & $56.9$ \\
    BLINK single-image (6 tasks) & 793 & 41.9 & $41.7$ & $50.9$ & $41.2$ & $56.6$ \\
    BLINK multi-image (8 tasks)  & 1108 & 37.1 & $37.8$ & $39.6$ & $38.1$ & $46.9$ \\
    \bottomrule
  \end{tabular}
\end{table}

\begin{wraptable}{r}{0.48\textwidth}
  \centering
  \scriptsize
  \begin{tabular}{lccc}
    \toprule
    reader & MMMU-Pro & MMMU & BLINK6 \\
    \midrule
    4B & 31.9 & 49.1 & 44.4 \\
    8B & 41.3 & 63.0 & 47.0 \\
    14B & 45.3 & 62.6 & 49.8 \\
    32B & 43.4 & 64.2 & 52.6 \\
    \bottomrule
  \end{tabular}
  \caption{Qwen3 models after full recipe (SFT $+$ RL), accuracy \%.
  Reasoning mode on (\texttt{/think}), greedy decode, 6{,}144-token cap,
  official parser; parse fallback ${\le}2\%$ per cell. Pre-RL baselines
  under the identical protocol are in Appendix~\ref{app:training}.}
  \label{tab:rung-rl}
\end{wraptable}


\subsection{What vision improves and what it does not}
\label{sec:blinksplit}

\begin{figure}[!t]
  \centering
  \begin{tikzpicture}
    \begin{axis}[
        width=\textwidth, height=6.4cm,
        ymin=0, ymax=90, ylabel={Accuracy (\%)}, ymajorgrids,
        unbounded coords=jump,
        symbolic x coords={Counting, Spatial Rel., Rel. Depth,
          Sem. Corr., IQ Test, Vis. Corr., Art Style, Func. Corr.,
          Forensics, Jigsaw, Multi-view, Localization, Similarity,
          Reflectance},
        xtick=data, x tick label style={rotate=40, anchor=east, font=\scriptsize},
        legend style={at={(0.98,0.98)}, anchor=north east, font=\scriptsize},
        legend cell align=left]
      \addplot[mark=none, color=gray!45, line width=0.9pt, forget plot] coordinates {
        (Spatial Rel.,49.7) (Spatial Rel.,74.1) (Spatial Rel.,nan)
        (Rel. Depth,50.0) (Rel. Depth,70.2) (Rel. Depth,nan)
        (Counting,40.8) (Counting,52.5) (Counting,nan)
        (Localization,44.3) (Localization,48.4) (Localization,nan)
        (IQ Test,22.7) (IQ Test,34.0) (IQ Test,nan)
        (Reflectance,42.5) (Reflectance,28.4) (Reflectance,nan)
        (Art Style,50.4) (Art Style,56.4) (Art Style,nan)
        (Jigsaw,47.3) (Jigsaw,49.3) (Jigsaw,nan)
        (Multi-view,53.4) (Multi-view,48.9) (Multi-view,nan)
        (Similarity,51.1) (Similarity,45.2) (Similarity,nan)
        (Sem. Corr.,22.3) (Sem. Corr.,35.3) (Sem. Corr.,nan)
        (Vis. Corr.,27.9) (Vis. Corr.,32.6) (Vis. Corr.,nan)
        (Func. Corr.,30.0) (Func. Corr.,26.2) (Func. Corr.,nan)
        (Forensics,23.5) (Forensics,25.8)};
      \addplot[only marks, mark=-, mark size=5.5pt, color=gray!70, very thick] coordinates {
        (Spatial Rel.,50) (Rel. Depth,50) (Art Style,50)
        (Counting,25) (Jigsaw,50) (Multi-view,50) (Localization,50)
        (Similarity,50) (Sem. Corr.,25) (IQ Test,25) (Vis. Corr.,25)
        (Reflectance,33.3) (Func. Corr.,25) (Forensics,25)};
      \addplot[only marks, mark=*, mark size=3.2pt, mark options={fill=gray!45}, color=gray!40!black, thick] coordinates {
        (Spatial Rel.,49.7) (Rel. Depth,50.0) (Counting,40.8)
        (Localization,44.3) (IQ Test,22.7) (Reflectance,42.5)
        (Art Style,50.4) (Jigsaw,47.3) (Multi-view,53.4) (Similarity,51.1)
        (Sem. Corr.,22.3) (Vis. Corr.,27.9) (Func. Corr.,30.0)
        (Forensics,23.5)};
      \addplot[only marks, mark=square*, mark size=3.0pt, mark options={fill=orange!80!black}, color=orange!80!black] coordinates {
        (Spatial Rel.,74.1) (Rel. Depth,70.2) (Counting,52.5)
        (Localization,48.4) (IQ Test,34.0) (Reflectance,28.4)};
      \addplot[only marks, mark=diamond*, mark size=3.8pt, mark options={fill=orange!40}, color=orange!80!black, thick] coordinates {
        (Art Style,56.4) (Jigsaw,49.3) (Multi-view,48.9) (Similarity,45.2)
        (Sem. Corr.,35.3) (Vis. Corr.,32.6) (Func. Corr.,26.2)
        (Forensics,25.8)};
      \legend{chance, blind, adapter (single-image), adapter (multi-image)}
    \end{axis}
  \end{tikzpicture}
  \caption{\textbf{Specific vision capabilities for GLM-5.2 adapted to vision.} All
  14 BLINK tasks for GLM-5.2 system, ordered by margin over chance. Squares indicate single-image tasks, and diamonds multi-image tasks. Decode configuration as in Section~\ref{sec:formulation},
  identical for vision and blind arms.}
  \label{fig:demand}
\end{figure}
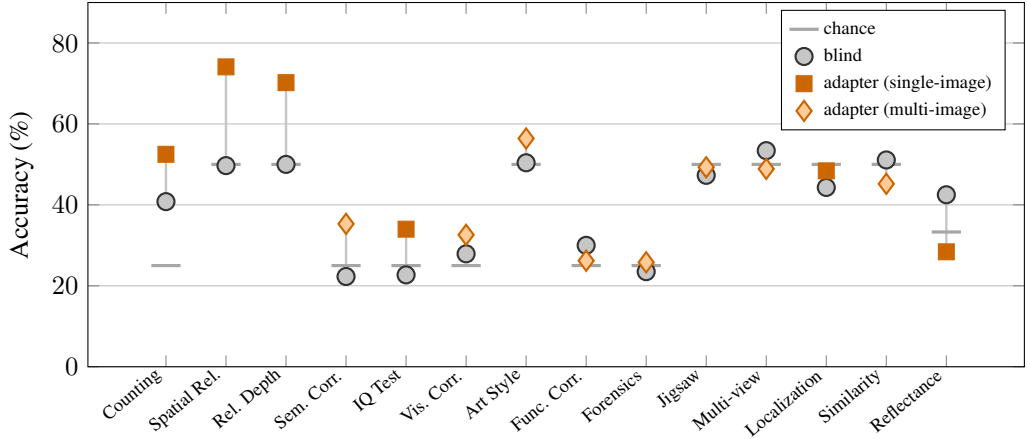

Vision's contribution to GLM-5.2 is concentrated, not uniform
(Figure~\ref{fig:demand}; all numbers in this subsection are the GLM-5.2
system). On the six single-image tasks, vision-minus-blind ranges from
$+24.4$ (Spatial Relation) and $+20.2$ (Relative Depth) down to $+4.1$
(Object Localization); the one negative is Relative Reflectance ($-14.1$),
where the blind score is pure text prior and reasoning over the image talks
the model out of it. The override is reader-specific: on the same task the
post-RL Qwen3-32B system holds its prior level (41.0 vs GLM-5.2's 28.4),
a concrete case of visual information arriving but being overridden by the
prior. The eight multi-image tasks tell the opposite story:
images buy $+1.8$ points overall (39.6\% vs 37.8\% blind), with only
Semantic Correspondence clearly above blind ($+13.0$) and two tasks below
it. Nearly half of the multi-image responses (44.5\%) never produce a
parseable answer, against 1.7\% blind on the same questions: the failure
is upstream of accuracy, and it is the images that cause it. The system was
trained exclusively on single-image data; cross-image comparison does not
emerge from single-image training.

\paragraph{Evaluation provenance.}
All numbers in this section come from serving the released artifacts on
their own runtimes and scoring with the official evaluator; decode
parameters, per-task fallback rates, and raw responses are archived per
row.

Per-subject MMMU deltas tell the same story at finer grain: vision helps
where the image carries data that the question does not restate (Accounting:
blind 36.7\%, vision 56.7\%), and does nothing where the question is
answerable from knowledge (Art: blind 63.3\%, vision 56.7\%).

\subsection{Reliability under controlled corruptions}
\label{sec:robustness}

The GLM-5.2 adapter gains $14.5$ points over its blind control on MMMU-Pro ($56.8\%$ vs.\ $42.3\%$; Table~\ref{tab:main}). A repeat paired control for an additional GLM-5.3 adapter trained with the GLM-5.2 recipe was completed. Under its separately frozen generation and scoring contract, GLM-5.3 gains $21.04$ points on MMMU-Pro ($62.89\%$ vs.\ $41.85\%$) establishing a similar image dependence within GLM-5.3. The moderate noise, brightness, and saturation grid does not show a clear aggregate loss. A separate 180-row panel selected severe levels for confirmation. On the remaining 1{,}550 MMMU-Pro rows, noise $\sigma{=}1.0$ and brightness $12\times$ reduce accuracy by $21.42$ and $17.48$ points, respectively, with familywise 95\% intervals excluding zero. Appendix~\ref{app:glm53-robustness} reports intervals, the full grid, a representative transformed image (Figure~\ref{fig:glm53-corruption-atlas}), selection rules, and scoring details.

\subsection{Trainable vision capabilities}
As the trainable parameter dimension is relatively small, we can hope to directly train on specific vision capabilities with smaller datasets with less risk of overfitting. See Table~\ref{tab:capability-matrix} for results finetuning on specific BLINK categories, using the training recipe set up by LIVR. While some visual categories are more fine-tuneable than others (e.g. visual correspondence and similarity), all finetuning did not degrade performance on held-out tasks. Using a smaller-scale language model still allows tuneability, but some of the correspondence tasks like semantic and functional correspondence are not as tuneable with a smaller language model component (see Appendix~\ref{app:capability-comparison}).
\begin{table*}[tbp]
  \centering
  \caption{\textbf{Task-specific projector training in GLM-5.2.}
Each row trains only the 49.5M-parameter projector on 1{,}000
LIVR-constructed examples, keeping GLM-5.2 and MoonViT frozen.
Training uses 10 epochs, a learning rate of $10^{-4}$, and
one seed. We report final-checkpoint accuracy (\%) on the
LIVR holdout, the corresponding BLINK task, and the remaining
13 BLINK tasks pooled together ($n=1{,}778$).
\textit{Base} and \textit{adapt} denote scores before and
after training. BLINK evaluation uses teacher-forced
choice-likelihood scoring on images composited into a
single input image.
$\dagger$: Evaluation on held-out HPatches and FunKPoint splits;
see LIVR~\citep{li2025livr}.}
  \label{tab:capability-matrix}
  \begin{tabular}{l rrr rrr rrr}
    \toprule
    & \multicolumn{3}{c}{LIVR holdout} & \multicolumn{3}{c}{BLINK: trained task}
      & \multicolumn{3}{c}{BLINK: other 13 tasks} \\
    \cmidrule(lr){2-4} \cmidrule(lr){5-7} \cmidrule(lr){8-10}
    Capability & base & adapt & $\Delta$ & base & adapt & $\Delta$
      & base & adapt & $\Delta$ \\
    \midrule
    Visual corr.$^\dagger$   & 19.2 & 64.0 & $+44.8$ & 26.2 & 56.4 & $+30.2$ & 41.8 & 45.5 & $+3.6$ \\
    Visual similarity         & 60.8 & 89.2 & $+28.4$ & 57.8 & 84.4 & $+26.7$ & 39.1 & 45.0 & $+5.9$ \\
    Art style                 & 51.2 & 81.6 & $+30.4$ & 53.8 & 76.9 & $+23.1$ & 39.5 & 42.1 & $+2.5$ \\
    Semantic corr.            & 26.0 & 45.6 & $+19.6$ & 23.7 & 40.3 & $+16.5$ & 41.7 & 49.2 & $+7.4$ \\
    Object localization & 56.4 & 89.2 & $+32.8$ & 47.5 & 63.1 & $+15.6$ & 39.3 & 43.4 & $+4.1$ \\
    Functional corr.$^\dagger$ & 22.2 & 19.4 & $-2.8$  & 23.8 & 33.8 & $+10.0$ & 41.6 & 44.2 & $+2.6$ \\
    Relative reflectance      & 47.2 & 50.8 & $+3.6$  & 27.6 & 35.8 & $+8.2$  & 41.4 & 46.1 & $+4.8$ \\
    Jigsaw                    & 42.4 & 81.6 & $+39.2$ & 50.7 & 56.7 & $+6.0$  & 39.5 & 39.3 & $-0.2$ \\
    Counting                  & ---  & ---  & ---     & 40.0 & 45.8 & $+5.8$  & 40.4 & 43.3 & $+2.8$ \\
    \bottomrule
  \end{tabular}
\end{table*}
\raggedbottom
\section{Inside the frozen reader: a Jacobian lens on the adapter's output}
\label{sec:interp}

We analyze visual representations in the released GLM-5.2 model with its trained projector and test how changes to image-token activations affect its predictions. At 743B parameters the language model is larger than those examined in prior lens studies of vision-language models. We compare the results with the smaller Qwen3 systems.

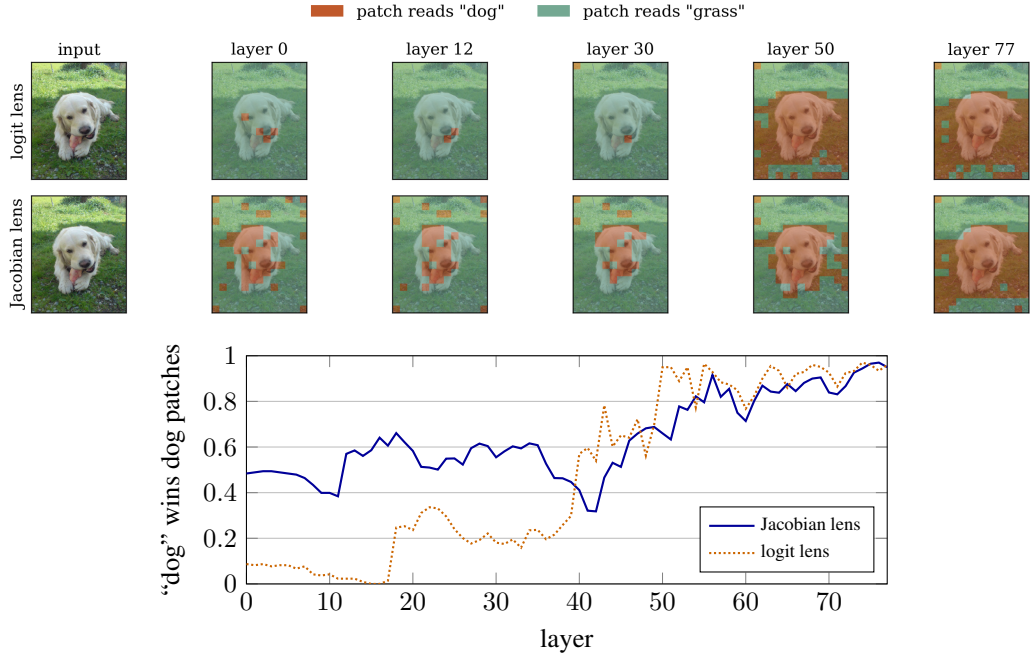
\begin{figure}
  \centering
  \includegraphics[width=0.98\textwidth]{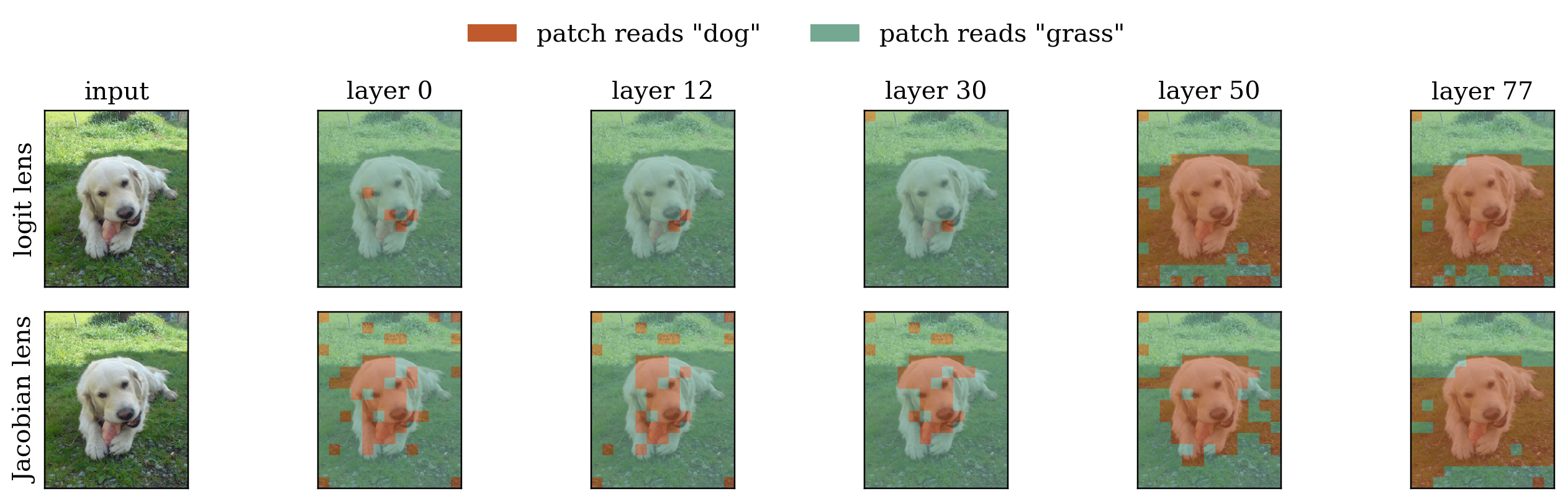}\\[4pt]
  \begin{tikzpicture}
    \begin{axis}[
        width=0.72\textwidth, height=4.6cm,
        xmin=0, xmax=77,
        xlabel={layer}, ylabel={``dog'' wins dog patches},
        ymin=0, ymax=1, ymajorgrids,
        legend style={at={(0.98,0.05)}, anchor=south east, font=\scriptsize},
        legend cell align=left]
      \addplot[blue!60!black, thick, mark=none] coordinates {
        (0,0.484) (1,0.489) (2,0.494) (3,0.494) (4,0.489) (5,0.484) (6,0.479) (7,0.464) (8,0.434) (9,0.399) (10,0.399) (11,0.384) (12,0.570) (13,0.585) (14,0.561) (15,0.586) (16,0.641) (17,0.606) (18,0.661) (19,0.621) (20,0.583) (21,0.513) (22,0.510) (23,0.501) (24,0.549) (25,0.550) (26,0.523) (27,0.595) (28,0.615) (29,0.604) (30,0.555) (31,0.581) (32,0.603) (33,0.594) (34,0.616) (35,0.608) (36,0.527) (37,0.464) (38,0.463) (39,0.447) (40,0.411) (41,0.321) (42,0.318) (43,0.466) (44,0.531) (45,0.513) (46,0.628) (47,0.659) (48,0.682) (49,0.688) (50,0.660) (51,0.633) (52,0.778) (53,0.763) (54,0.822) (55,0.796) (56,0.916) (57,0.820) (58,0.855) (59,0.750) (60,0.714) (61,0.801) (62,0.869) (63,0.843) (64,0.838) (65,0.876) (66,0.845) (67,0.881) (68,0.900) (69,0.905) (70,0.839) (71,0.831) (72,0.867) (73,0.926) (74,0.944) (75,0.965) (76,0.970) (77,0.950)};
      \addplot[orange!80!black, thick, densely dotted, mark=none] coordinates {
        (0,0.087) (1,0.082) (2,0.087) (3,0.077) (4,0.082) (5,0.082) (6,0.067) (7,0.077) (8,0.042) (9,0.037) (10,0.042) (11,0.023) (12,0.023) (13,0.023) (14,0.009) (15,0.000) (16,0.000) (17,0.014) (18,0.247) (19,0.253) (20,0.236) (21,0.311) (22,0.336) (23,0.331) (24,0.296) (25,0.240) (26,0.201) (27,0.177) (28,0.192) (29,0.221) (30,0.181) (31,0.175) (32,0.195) (33,0.159) (34,0.235) (35,0.237) (36,0.196) (37,0.216) (38,0.258) (39,0.298) (40,0.570) (41,0.596) (42,0.540) (43,0.783) (44,0.602) (45,0.651) (46,0.641) (47,0.722) (48,0.562) (49,0.699) (50,0.950) (51,0.950) (52,0.889) (53,0.950) (54,0.773) (55,0.965) (56,0.929) (57,0.884) (58,0.874) (59,0.849) (60,0.768) (61,0.823) (62,0.900) (63,0.955) (64,0.934) (65,0.860) (66,0.920) (67,0.929) (68,0.960) (69,0.951) (70,0.926) (71,0.865) (72,0.924) (73,0.933) (74,0.970) (75,0.960) (76,0.934) (77,0.960)};
      \legend{Jacobian lens, logit lens}
    \end{axis}
  \end{tikzpicture}
  \caption{\textbf{The logit lens reads ``grass'' everywhere early whereas
  the Jacobian lens sees the dog from the first layers.} Top rows: per-patch
  winner between ``dog'' and ``grass'' overlaid on the image at five depths
  (GLM, 743B) for each image token. Bottom: fraction of dog patches read as
  ``dog'' per layer (grass patches stay ``grass'' at 0.7--0.9 under the
  Jacobian lens throughout). Grass labels use a color heuristic; the
  mask-based control is Appendix Figure~\ref{fig:interp-discrim}.}
  \label{fig:interp-grass}
\end{figure}

\paragraph{Where does the vision work happen?}
There are two hypotheses. Under (A), the frozen GLM reader does most of the work: it receives surface-level features from the encoder and reshapes them into concepts over its layers. Under (B), the vision side, the 417M frozen encoder plus the 50M adapter that is the only trained part, already delivers the concepts, and the reader only translates them into its own vocabulary. A logit lens, which scores each layer's hidden state against the output layer's word embeddings, appears to support (A): on a dog-on-grass image, ``dog'' begins to win the dog patches only in the middle layers (Figure~\ref{fig:interp-grass}, top row), as
\citet{neo2025towards} also report for LLaVA. But the logit lens reads every layer in the output layer's word coordinates, and so inherits the reader's word priors (Appendix~\ref{app:interp-winner},
\ref{app:interp-control}). We therefore use a Jacobian lens \citep{gurnee2026workspace}: each word's unembedding row is pulled back through the remaining layers by differentiating the word's final logit with
respect to the hidden state at layer $\ell$ (Appendix~\ref{app:interp-winner}). This accounts, to first order, for everything the remaining layers do to the hidden state, including rotation, attention mixing, and MLPs. Under the Jacobian lens the picture reverses: the dog patches read as ``dog'' from layer 12, while the grass patches still read as ``grass'' (Figure~\ref{fig:interp-grass}, bottom row). A mask-based control confirms this (Appendix Figure~\ref{fig:interp-discrim}), and a linear probe on the adapter's output separates object from background at 0.96 AUC, with all 78 reader layers together adding at most 0.03. These results support (B): the concepts arrive from the adapter, and the reader's depth translates them into the vocabulary basis. Geometrically, the image tokens merge into the text tokens' subspace as depth increases: cosine between the mean image token and the mean text token is 0.02 at the input, 0.46 by layer 12, and 0.91 by layer 77 (Appendix~\ref{app:interp-geometry}, Figure~\ref{fig:interp-gap-pca}). Hyperparameter details and further findings are in Appendix~\ref{app:steering}.

\FloatBarrier
\newpage
\flushbottom
\section{Related Work}
\label{sec:related}

\subsection{Small adapters between frozen vision/language models}

Bridging a frozen language model with a small trained module is an old
recipe~\citep{tsimpoukelli2021frozen, merullo2023limber, li2023blip2, zhu2023minigpt4, koh2023fromage, shukor2023epalm, alayrac2022flamingo}. These systems have  stalled on hard benchmarks (BLIP-2 and
MiniGPT-4 score 34.0\% and 27.6\% on MMMU \citep{yue2024mmmu}), and the
accepted explanation was projector capacity~\citep{dai2024nvlm}. The notable difference between our work and previous adapter-based training methods is the scale of the language model, namely 743B for GLM 5.2. Previous work has studied the idea that models' learned representations get more similar as the scale of the model and datasets grow~\citep{huh2024platonic,moschella2023relative,
maiorca2023latent}.

\subsection{Reinforcement learning to train vision-to-text adapters}
Reinforcement learning has been used to train visual adapters by
  rewarding the outputs they elicit from a frozen language
  model~\citep{yu2023esper,skywork2025r1v2}. We do not propose a novel training method, but rather a recipe for RL training vision models at scale, and a study on how the language model scale affects visual capabilities.

\subsection{Interpretability of vision-to-text adapters}
Previous papers have also studied the impact of individual visual tokens in the multimodal model's answers through mechanistic interpretability tools, such as corrected versions of logit lens~\citep{krojer2026latentlens}. We provide this interpretability analysis for our tuned adapter model as a complement to the concrete visual baselines.

\section{Discussion \& Limitations}
\label{sec:discussion}

\paragraph{Visual capability scales with the frozen reader.}
With the encoder, adapter architecture, data, and recipe held fixed, benchmark accuracy tracks the size of the language model. After the full recipe, MMMU-Pro rises from $31.9$ to $41.3$ to $45.3$ across the Qwen3 4B, 8B, and 14B models (Table~\ref{tab:rung-rl}), plateaus at 32B ($43.4$, within the confidence interval of 14B), and is about 12 points higher for the two frontier models, which land within $0.1$ points of each other ($56.8$ vs $56.9$). Accuracy without the image stays flat across all sizes (Figure~\ref{fig:scalinglaw}). The contribution of the RL stage shrinks as the language model grows. What RL provides is a reliable return to the reasoning mode that SFT with empty reasoning blocks suppressed, and larger models return to it from the prompt switch alone. The adapter is $\approx$0.007\% of the largest model's parameters, so the gains from scale come almost entirely from the frozen language model rather than from anything the adapter learns differently at each size.


\paragraph{Robustness to image corruption is bounded.}
On MMMU-Pro, GLM-5.3 depends materially on the input image. Moderate noise, brightness, and saturation changes produce no clear loss in accuracy, but severe noise and brightening reduce accuracy by $21.42$ and $17.48$ points on a confirmation set disjoint from the rows used to select those levels. This does not establish invariance to corruption in general, nor a smooth relationship between corruption strength and accuracy. It shows that the system tolerates moderate corruptions and fails under severe ones.

\paragraph{Limitations.}
We test one encoder family and one SFT corpus, and the alignment residuals of Eq.~\ref{eq:align} are measured for two language-model families only. The main runs are single seeds, and we do not measure seed variance. Because the language model is frozen, behaviors conveyed through the adapter act like instructions in the prompt: a stronger prompt in the same channel can override them, so any behavioral guarantee trained into the adapter is weak. Rewards and answer equivalence rely on a pinned LLM judge,
and all benchmarks are multiple-choice, so open-ended grounded use is not measured.

\section{Conclusion}
Training $\approx$50M adapter parameters is enough to give a frozen text-only language model strong visual competence ($56.8\%$ on MMMU-Pro and $71.4\%$ on MMMU at 743B), with near-identical results for two unrelated frontier models. Single-image perception saturates early in model size, and abilities absent from the training data, such as cross-image comparison, do not emerge at any size. Inside the frozen model, the adapter's output is readable in gradient coordinates from the first layers; it enters nearly orthogonal to the text tokens and merges with them over depth. Because the language model is frozen, the same setup is both a practical route to vision for deployed text-only models and a controlled instrument for studying how visual capability arrives.


\bibliographystyle{plainnat}
\bibliography{ref}


\appendix
\clearpage
\section{Training details}
\label{app:training}

This appendix collects the SFT and RL constants shared by every Qwen3 run, the parse-fallback rates behind Table~\ref{tab:main}, the training curves for all readers, the pre-RL baselines, and derivations for the embedding-alignment initialization and the RL objective.

\begin{table}[htbp]
  \caption{Supervised fine-tuning constants (identical at every scale).}
  \label{tab:sft-constants}
  \centering
  \small
  \begin{tabular}{ll}
    \toprule
    global batch & 128 conversations \\
    optimizer & AdamW, $\beta = (0.9, 0.95)$, weight decay $0.1$ \\
    learning rate & $3 \times 10^{-4}$, constant after 50-step linear warmup \\
    gradient clip & 1.0 \\
    precision / max sequence length & bf16 / 5{,}120 tokens (padded) \\
    steps & 2{,}048 (2 epochs) \\
    shuffle & single fixed-seed shuffle, shared across runs \\
    \bottomrule
  \end{tabular}
\end{table}

\begin{table}[htbp]
  \caption{Reinforcement-learning constants (identical at every scale).}
  \label{tab:rl-constants}
  \centering
  \small
  \begin{tabular}{ll}
    \toprule
    prompts / group size per update & 64 / 8 (512 responses) \\
    sampling & temperature 1.0, top-$p$ 0.95 \\
    response cap & 6{,}144 tokens \\
    clip / KL & $\varepsilon_{\mathrm{high}} = 0.28$ / $\beta = 10^{-3}$ \\
    learning rate & $1 \times 10^{-5}$, projector only \\
    updates / checkpointing & 150 / every update \\
    correctness judge & LLM judge against reference answers; judge failure aborts the update \\ 
    actor--rollout parity gate & actor and rollout log-probs agree: mean $|\Delta \log p| \leq 0.05$ nats/token3 \\
    \bottomrule
  \end{tabular}
\end{table}

\begin{table}[htbp]
  \caption{Pre-RL baselines: the SFT-only adapter, evaluated with the
  reader's reasoning mode on (\texttt{/think}) under the protocol of
  Table~\ref{tab:rung-rl}. Accuracy \% (in parentheses: \% of responses
  with a non-empty reasoning block).}
  \label{tab:sft-think-baselines}
  \centering
  \small
  \begin{tabular}{lccc}
    \toprule
    reader & MMMU-Pro & MMMU & BLINK6 \\
    \midrule
    4B & 26.5 (19) & 47.6 (11) & 43.4 (0) \\
    8B & 35.6 (26) & 51.8 (18) & 49.6 (8) \\
    14B & 44.5 (81) & 60.5 (79) & 49.4 (58) \\
    32B & 44.4 (62) & 64.1 (72) & 49.9 (80) \\
    \bottomrule
  \end{tabular}
\end{table}

\begin{table}[htbp]
  \caption{Parse-fallback rates (\%) for the arms of Table~\ref{tab:main}:
  the share of responses with no parseable answer, scored by seeded random
  choice.}
  \label{tab:fallback}
  \centering
  \small
  \begin{tabular}{lcccc}
    \toprule
    & \multicolumn{2}{c}{GLM-5.2} & \multicolumn{2}{c}{Nemotron-3-Ultra} \\
    \cmidrule(lr){2-3}\cmidrule(lr){4-5}
    & Blind & Vision & Blind & Vision \\
    \midrule
    MMMU (standard, val)         & ${\le}3$ & 10.1 & ${\le}3$ & 10.8 \\
    MMMU-Pro (standard)          & ${\le}3$ & ${\le}3$ & 5.0 & ${\le}3$ \\
    BLINK single-image (6 tasks) & ${\le}3$ & ${\le}3$ & ${\le}3$ & 16.8 \\
    BLINK multi-image (8 tasks)  & ${\le}3$ & 44.5 & ${\le}3$ & 30.0 \\
    \bottomrule
  \end{tabular}
\end{table}

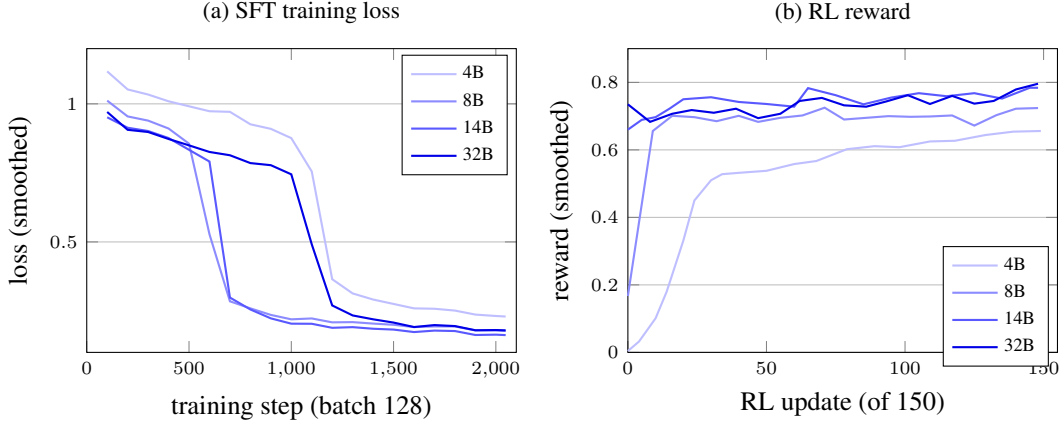
\begin{figure}[htbp]
  \centering
  \begin{tikzpicture}
    \begin{groupplot}[
        group style={group size=2 by 1, horizontal sep=42pt},
        width=0.52\textwidth, height=5.6cm,
        tick label style={font=\scriptsize},
        legend style={font=\scriptsize}, legend cell align=left]
      \nextgroupplot[title={\small (a) SFT training loss},
        xmin=0, xmax=2100, ymin=0.1, ymax=1.2, ymajorgrids,
        xlabel={training step (batch 128)}, ylabel={loss (smoothed)},
        legend style={at={(0.98,0.98)}, anchor=north east}]
      \addplot[blue!25, thick, mark=none] coordinates {
        (100,1.1176) (200,1.0522) (300,1.0338) (400,1.0093) (500,0.9912) (600,0.9733) (700,0.9711) (800,0.9260) (900,0.9095) (1000,0.8759) (1100,0.7551) (1200,0.3650) (1300,0.3135) (1400,0.2914) (1500,0.2754) (1600,0.2594) (1700,0.2580) (1800,0.2514) (1900,0.2369) (2000,0.2318) (2048,0.2299)};
      \addplot[blue!45, thick, mark=none] coordinates {
        (100,1.0117) (200,0.9545) (300,0.9393) (400,0.9107) (500,0.8558) (600,0.5279) (700,0.2850) (800,0.2594) (900,0.2357) (1000,0.2197) (1100,0.2229) (1200,0.2088) (1300,0.2096) (1400,0.2043) (1500,0.2000) (1600,0.1916) (1700,0.1939) (1800,0.1939) (1900,0.1795) (2000,0.1797) (2048,0.1775)};
      \addplot[blue!65, thick, mark=none] coordinates {
        (100,0.9513) (200,0.9135) (300,0.9016) (400,0.8762) (500,0.8336) (600,0.7911) (700,0.2986) (800,0.2554) (900,0.2232) (1000,0.2040) (1100,0.2036) (1200,0.1890) (1300,0.1914) (1400,0.1856) (1500,0.1826) (1600,0.1736) (1700,0.1789) (1800,0.1773) (1900,0.1630) (2000,0.1642) (2048,0.1626)};
      \addplot[blue!90!black, thick, mark=none] coordinates {
        (100,0.9706) (200,0.9060) (300,0.8981) (400,0.8728) (500,0.8497) (600,0.8258) (700,0.8139) (800,0.7855) (900,0.7779) (1000,0.7448) (1100,0.4914) (1200,0.2700) (1300,0.2338) (1400,0.2198) (1500,0.2079) (1600,0.1918) (1700,0.1989) (1800,0.1953) (1900,0.1801) (2000,0.1809) (2048,0.1798)};
      \legend{4B, 8B, 14B, 32B}
      \nextgroupplot[title={\small (b) RL reward},
        xmin=0, xmax=155, ymin=0, ymax=0.9, ymajorgrids,
        xlabel={RL update (of 150)}, ylabel={reward (smoothed)},
        legend style={at={(0.98,0.35)}, anchor=north east}]
      \addplot[blue!25, thick, mark=none] coordinates {
        (0,0.004) (4,0.032) (10,0.101) (14,0.180) (20,0.331) (24,0.450) (30,0.510) (34,0.528) (40,0.532) (50,0.538) (60,0.558) (68,0.567) (79,0.602) (89,0.611) (98,0.608) (109,0.625) (118,0.627) (129,0.644) (139,0.654) (149,0.656)};
      \addplot[blue!45, thick, mark=none] coordinates {
        (0,0.167) (9,0.656) (16,0.701) (24,0.697) (32,0.685) (40,0.701) (47,0.683) (55,0.695) (63,0.702) (71,0.725) (78,0.690) (86,0.695) (94,0.700) (102,0.698) (109,0.699) (117,0.702) (125,0.672) (133,0.703) (140,0.722) (148,0.724)};
      \addplot[blue!65, thick, mark=none] coordinates {
        (0,0.660) (5,0.689) (10,0.697) (15,0.722) (20,0.750) (30,0.756) (40,0.742) (50,0.736) (60,0.728) (65,0.783) (75,0.762) (85,0.735) (95,0.755) (105,0.768) (115,0.759) (125,0.768) (135,0.752) (145,0.784) (148,0.784)};
      \addplot[blue!90!black, thick, mark=none] coordinates {
        (0,0.735) (8,0.683) (16,0.707) (23,0.718) (31,0.710) (39,0.722) (47,0.694) (55,0.707) (62,0.745) (70,0.754) (78,0.732) (86,0.728) (93,0.743) (101,0.762) (109,0.736) (117,0.760) (125,0.737) (132,0.745) (140,0.779) (148,0.796)};
      \legend{4B, 8B, 14B, 32B}
    \end{groupplot}
  \end{tikzpicture}
  \caption{\textbf{Training across reader scales.} (a) SFT training loss
  (final losses 4B $0.230$, 8B $0.178$, 14B $0.163$, 32B $0.180$;
  comparable within the shared-tokenizer Qwen3 family only). Every scale
  shows a grokking-like delayed transition: an extended plateau followed by
  an abrupt drop (Appendix~\ref{sec:ablations}). (b) RL reward on rendered
  math (Section~\ref{sec:method-rl}), same recipe at every scale; all four
  runs complete 150 updates. Initial reward tracks reader scale (4B starts
  at zero, 32B at $0.72$); the spread narrows from $0.72$ at update 0 to
  $0.14$ at update 150 (32B ends highest, $\approx$0.80).
  }
  \label{fig:losslaw}
\end{figure}

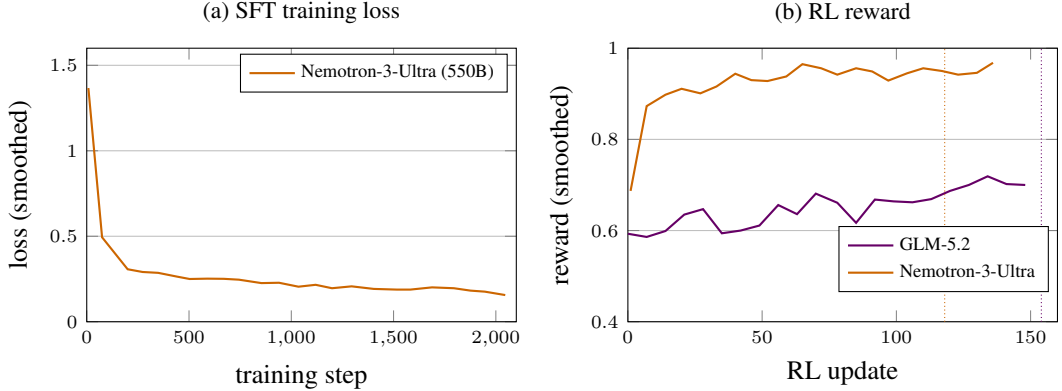
\begin{figure}[!t]
  \centering
  \begin{tikzpicture}
    \begin{groupplot}[
        group style={group size=2 by 1, horizontal sep=42pt},
        width=0.52\textwidth, height=5.2cm,
        tick label style={font=\scriptsize},
        legend style={font=\scriptsize}, legend cell align=left]
      \nextgroupplot[title={\small (a) SFT training loss},
        xmin=0, xmax=2100, ymin=0, ymax=1.6, ymajorgrids,
        xlabel={training step}, ylabel={loss (smoothed)},
        legend style={at={(0.98,0.98)}, anchor=north east}]
            \addplot[orange!80!black, thick, mark=none] coordinates {
        (8,1.367) (75,0.494) (200,0.307) (273,0.291) (349,0.286) (441,0.264) (503,0.250) (589,0.252) (670,0.251) (744,0.246) (854,0.226) (940,0.228) (1035,0.205) (1119,0.216) (1199,0.196) (1295,0.207) (1403,0.192) (1516,0.188) (1582,0.188) (1688,0.201) (1795,0.196) (1876,0.182) (1946,0.176) (2046,0.156)};
      \legend{Nemotron-3-Ultra (550B)}
      \nextgroupplot[title={\small (b) RL reward},
        xmin=0, xmax=160, ymin=0.4, ymax=1.0, ymajorgrids,
        xlabel={RL update}, ylabel={reward (smoothed)},
        legend style={at={(0.98,0.35)}, anchor=north east}]
      \addplot[violet!80!black, thick, mark=none] coordinates {
        (0,0.593) (7,0.586) (14,0.599) (21,0.635) (28,0.647) (35,0.594) (42,0.600) (49,0.611) (56,0.656) (63,0.636) (70,0.681) (78,0.661) (85,0.617) (92,0.668) (99,0.664) (106,0.662) (113,0.669) (120,0.687) (127,0.700) (134,0.719) (141,0.702) (148,0.700)};
      \addplot[orange!80!black, thick, mark=none] coordinates {
        (1,0.687) (7,0.873) (14,0.898) (20,0.911) (27,0.901) (33,0.916) (40,0.944) (46,0.930) (52,0.928) (59,0.938) (65,0.965) (72,0.956) (78,0.942) (85,0.956) (91,0.949) (97,0.929) (104,0.945) (110,0.956) (117,0.950) (123,0.942) (130,0.946) (136,0.968)};
      \draw[densely dotted, violet!80!black] (axis cs:154,0.4) -- (axis cs:154,1.0);
      \draw[densely dotted, orange!80!black] (axis cs:118,0.4) -- (axis cs:118,1.0);
      \legend{GLM-5.2, Nemotron-3-Ultra}
    \end{groupplot}
  \end{tikzpicture}
  \caption{\textbf{MoE readers' training curves} (the runs that produced
  the released checkpoints). (a) SFT loss: Nemotron-3-Ultra descends
  without a visible plateau. GLM-5.2's SFT log is too short to plot at this
  scale; its affine-initialized loss starts at $0.86$, already below the
  Qwen3 plateau levels, and descends smoothly. The plateau-then-collapse
  transition is a Qwen3-family observation. (b) RL reward on rendered math;
  dotted lines mark the released checkpoints (update $154$ for GLM-5.2,
  $118$ for Nemotron-3-Ultra). Loss and reward values are not comparable
  across readers or to the Qwen3 figures: the readers use different
  tokenizers, trainers, reward compositions, and judges. Only the curve
  shapes are comparable.}
  \label{fig:moe-training}
\end{figure}
\FloatBarrier

\subsection{Embedding-alignment initialization}
\label{app:align}

The source reader and the target reader share the tokenizer vocabulary
$\mathcal{T}$ but not the embedding space:
$e^{\mathrm{tgt}}_t \in \mathbb{R}^{d_{\mathrm{tgt}}}$ and
$e^{\mathrm{src}}_t \in \mathbb{R}^{d_{\mathrm{src}}}$. If a map
$g : \mathbb{R}^{d_{\mathrm{src}}} \to \mathbb{R}^{d_{\mathrm{tgt}}}$
satisfies $g(e_t^{\mathrm{src}}) = e_t^{\mathrm{tgt}}$ for all
$t \in \mathcal{T}$, then $g \circ P_{\mathrm{src}}$ is a valid projector
for the target reader. We restrict $g$ to the affine class $g(v) = Wv + c$
for two reasons. First, the fit is then ordinary least squares over the
shared tokens (Eq.~\ref{eq:align}), with a closed-form solution (the
regression of $e^{\mathrm{tgt}}$ on $e^{\mathrm{src}}$). Second, because
the source projector's output layer is itself linear,
$P_{\mathrm{src}}(v) = W_2\, h(v) + b_2$, the composition folds exactly into
the existing architecture with no new parameters:
\begin{equation}
  (g \circ P_{\mathrm{src}})(v)
  \;=\;
  W^{*}\bigl(W_2\, h(v) + b_2\bigr) + c^{*}
  \;=\;
  \underbrace{(W^{*} W_2)}_{W_2'}\, h(v)
  + \underbrace{(W^{*} b_2 + c^{*})}_{b_2'}.
  \label{eq:fold}
\end{equation}

\subsection{Reinforcement-learning objective}
\label{app:cispo}

With group-normalized advantages $\hat{A}_i$ (Section~\ref{sec:method-rl}),
the CISPO objective \citep{minimax2025m1} is
\begin{equation}
  J(\theta)
  =
  \mathbb{E}\!\left[
    \frac{1}{\sum_i |y_i|}
    \sum_{i=1}^{G} \sum_{t=1}^{|y_i|}
    \mathrm{sg}\!\bigl(\min(r_{i,t},\, 1 + \varepsilon_{\mathrm{high}})\bigr)\,
    \hat{A}_i \,
    \log \pi_{\theta}(y_{i,t} \mid y_{i,<t}, x_i)
  \right],
  \label{eq:cispo}
\end{equation}
where
$r_{i,t} = \pi_{\theta}(y_{i,t} \mid y_{i,<t}, x_i) /
\pi_{\theta_{\mathrm{old}}}(y_{i,t} \mid y_{i,<t}, x_i)$
is the token importance ratio between the updated and sampling policies,
clipped from above only and held out of the gradient ($\mathrm{sg}$). A KL
penalty against the sampling policy with coefficient $\beta$ is added to
this objective. 
Constants are in Table~\ref{tab:rl-constants}.

\clearpage
\section{Task-specific visual training for different language model bases}
\label{app:capability-comparison}
To better understand the role of the language model scale for specific task tuneability, we present comparative results with Qwen3-32B as the language model in 
Table~\ref{tab:capability-reader}. Both methods are able to tune to specific tasks with minimal forgetting on baseline tasks, and occasionally the value of the larger language model in GLM-5.2 stands out, namely adaptability in semantic correspondence and functional correspondence.

\begin{table}[htbp]
  \centering
  \small
  \setlength{\tabcolsep}{4pt}
  \caption{\textbf{Capability fine-tuning: GLM-5.2 versus Qwen3-32B.}
  Full version of Table~\ref{tab:capability-matrix}, to demonstrated the comparative utility of a larger language model.}
  \label{tab:capability-reader}
  \textbf{(a) BLINK: trained task}\par\smallskip
  \begin{tabular}{l rrr rrr}
    \toprule
    & \multicolumn{3}{c}{GLM-5.2} & \multicolumn{3}{c}{Qwen3-32B} \\
    \cmidrule(lr){2-4}\cmidrule(lr){5-7}
    Capability & base & adapt & $\Delta$ & base & adapt & $\Delta$ \\
    \midrule
    Visual corr. & 26.2 & 56.4 & $+30.2$ & 26.2 & 53.5 & $+27.3$ \\
    Visual similarity & 57.8 & 84.4 & $+26.7$ & 59.3 & 79.3 & $+20.0$ \\
    Art style & 53.8 & 76.9 & $+23.1$ & 47.0 & 70.9 & $+23.9$ \\
    Semantic corr. & 23.7 & 40.3 & $+16.5$ & 26.6 & 28.1 & $+1.4$ \\
    Object localization & 47.5 & 63.1 & $+15.6$ & 45.9 & 59.0 & $+13.1$ \\
    Functional corr. & 23.8 & 33.8 & $+10.0$ & 19.2 & 16.9 & $-2.3$ \\
    Relative reflectance & 27.6 & 35.8 & $+8.2$ & 29.9 & 41.8 & $+11.9$ \\
    Jigsaw & 50.7 & 56.7 & $+6.0$ & 46.7 & 57.3 & $+10.7$ \\
    Counting & 40.0 & 45.8 & $+5.8$ & 40.8 & 37.5 & $-3.3$ \\
    \bottomrule
  \end{tabular}

  \medskip
  \textbf{(b) LIVR holdout and BLINK: other 13 tasks}\par\smallskip
  \begin{tabular}{l rr rrr rrr}
    \toprule
    & \multicolumn{2}{c}{LIVR: adapt}
      & \multicolumn{3}{c}{GLM: other 13}
      & \multicolumn{3}{c}{Qwen: other 13} \\
    \cmidrule(lr){2-3}\cmidrule(lr){4-6}\cmidrule(lr){7-9}
    Capability & GLM & Qwen & base & adapt & $\Delta$
      & base & adapt & $\Delta$ \\
    \midrule
    Visual corr. & 64.0 & 59.6 & 41.8 & 45.5 & $+3.6$ & 41.4 & 41.4 & $+0.0$ \\
    Visual similarity & 89.2 & 88.8 & 39.1 & 45.0 & $+5.9$ & 38.6 & 40.9 & $+2.4$ \\
    Art style & 81.6 & 79.6 & 39.5 & 42.1 & $+2.5$ & 39.6 & 42.2 & $+2.6$ \\
    Semantic corr. & 45.6 & 34.8 & 41.7 & 49.2 & $+7.4$ & 41.1 & 42.5 & $+1.4$ \\
    Object localization & 89.2 & 85.6 & 39.3 & 43.4 & $+4.1$ & 39.6 & 43.6 & $+4.0$ \\
    Functional corr. & 19.4 & 21.5 & 41.6 & 44.2 & $+2.6$ & 41.6 & 42.5 & $+1.0$ \\
    Relative reflectance & 50.8 & 52.0 & 41.4 & 46.1 & $+4.8$ & 40.8 & 43.3 & $+2.5$ \\
    Jigsaw & 81.6 & 84.4 & 39.5 & 39.3 & $-0.2$ & 39.5 & 40.3 & $+0.8$ \\
    Counting & --- & --- & 40.4 & 43.3 & $+2.8$ & 40.0 & 38.8 & $-1.2$ \\
    \bottomrule
  \end{tabular}
\end{table}
\FloatBarrier

\clearpage
\section{GLM-5.3 MMMU-Pro robustness evaluation}
\label{app:glm53-robustness}

\paragraph{System.}
The GLM-5.3 system we tested uses the same frozen MoonViT encoder, approximately 50M-parameter projector architecture, training data, initialization, SFT, and RL recipe as the GLM-5.2 system. Only the projector is trained. Evaluation uses one frozen GLM-5.3 language-model revision and one projector checkpoint. 

\paragraph{Generation and scoring.}
All arms use temperature $1.0$, top-$p$ $0.95$, a $131{,}072$-token output cap, and generation seeds derived from a base seed of 20260904. Unlike the Table~\ref{tab:main} protocol, no reasoning-effort setting is passed. Accuracy is per row (each of the 1,730 MMMU-Pro items weighted equally), with answers extracted by the official MMMU-Pro parser at a pinned version. The no-image arm preserves the text and decoder settings but sends no image payload and removes the image placeholders from the prompt. No rows were excluded.

\begin{table}[htbp]
  \caption{GLM-5.3 accuracy (\%) with clean vision and a matched no-image
control. Intervals for clean minus no-image are 95\% subject-stratified
paired-bootstrap intervals based on 10,000 replicates.}

  \label{tab:glm53-image-dependence}
  \centering
  \small
  \begin{tabular}{lrrrrr}
    \toprule
    benchmark & $n$ & clean & no image & $\Delta$ (pp) & paired 95\% CI \\
    \midrule
    MMMU-Pro & 1{,}730 & 62.89 & 41.85 & +21.04 & [18.50, 23.64] \\
    \bottomrule
  \end{tabular}
\end{table}

\paragraph{Moderate corruption grid.}
Each corruption is applied independently to the original decoded 8-bit RGB image, before model-specific image normalization. Gaussian noise adds deterministically seeded i.i.d.\ zero-mean noise with $\sigma\in\{0.02,0.05,0.10\}$ to RGB values scaled $[0,1]$; values are then rounded and clipped to the 8-bit range. Brightness and saturation are adjusted using the \verb|ImageEnhance.Brightness| and \verb|ImageEnhance.Color| transformations respectively from the Pillow image-processing library. The brightness was obtained with factors $\{1.10,1.25,1.50\}$ and the saturation adjusted by factors $\{0.75,0.50,0.00\}$, where $1.00$ denotes the original image and $0.00$ produces a grayscale image. All transformed images are returned as 8-bit RGB. The same 1,730 MMMU-Pro rows in paired order are used for every arm. Table~\ref{tab:glm53-moderate-mmmu} reports accuracy under the official parser. No moderate condition differs clearly from clean; the noise and saturation deltas are all slightly positive, and with single-seed sampling at temperature $1.0$ shifts of this size are within run-to-run variation.

\begin{figure}[htbp]
  \centering
  \includegraphics[width=0.98\textwidth]{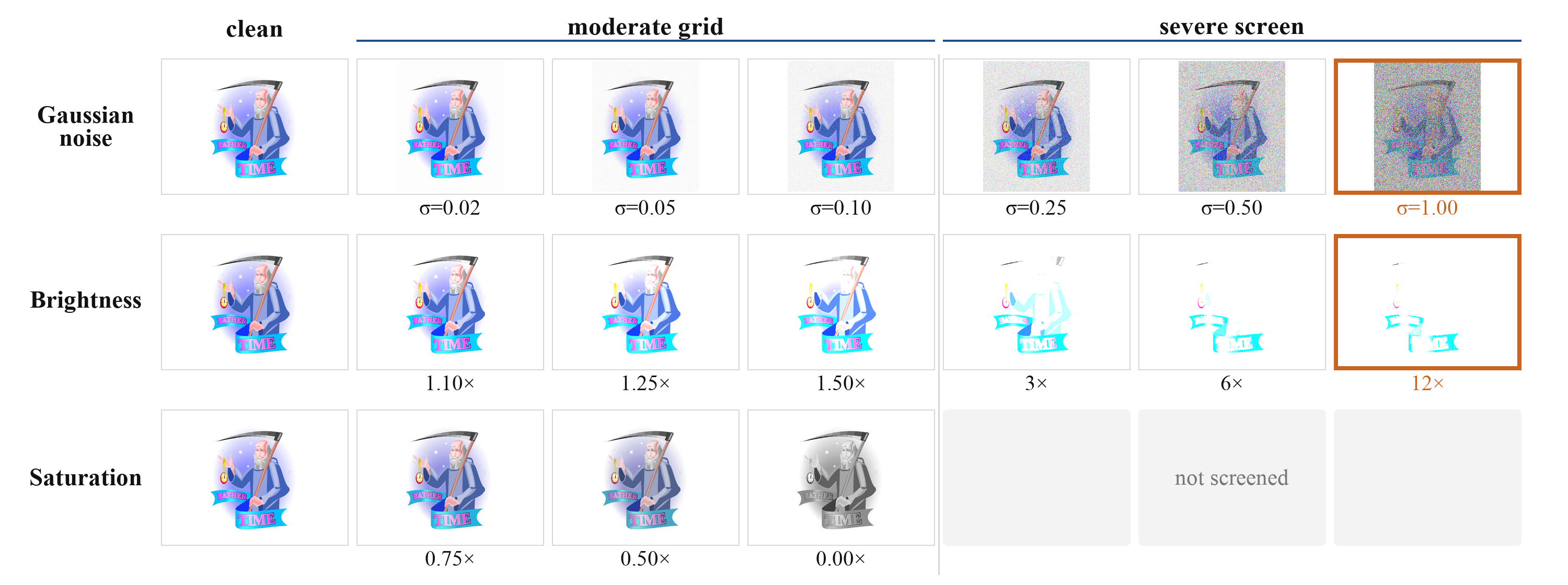}
  \caption{\textbf{Corruption families and levels.} One representative
  MMMU-Pro image transformed by the evaluation code. Noise $\sigma{=}0.10$
  and brightness $1.50\times$ occur in both the moderate grid and severe
  screen and are shown once under the moderate grid; the right columns are
  the additional screened levels. Orange frames mark the two levels carried
  into holdout confirmation. Saturation was not part of the severe screen.}
  \label{fig:glm53-corruption-atlas}
\end{figure}

\begin{table}[htbp]
  \caption{GLM-5.3 on the full MMMU-Pro moderate corruption grid. Changes and
  pointwise paired 95\% intervals are relative to the clean arm.}
  \label{tab:glm53-moderate-mmmu}
  \centering
  \small
  \begin{tabular}{lrrr}
    \toprule
    condition & accuracy (\%) & $\Delta$ vs.\ clean (pp) & paired 95\% CI \\
    \midrule
    clean & 62.89 & --- & --- \\
    noise $\sigma=0.02$ & 64.57 & $+1.68$ & [$-0.29$, 3.64] \\
    noise $\sigma=0.05$ & 64.74 & $+1.85$ & [$-0.12$, 3.82] \\
    noise $\sigma=0.10$ & 63.24 & $+0.35$ & [$-1.50$, 2.20] \\
    brightness $1.10\times$ & 62.60 & $-0.29$ & [$-2.25$, 1.68] \\
    brightness $1.25\times$ & 62.72 & $-0.17$ & [$-2.02$, 1.73] \\
    brightness $1.50\times$ & 62.20 & $-0.69$ & [$-2.66$, 1.27] \\
    saturation $0.75\times$ & 64.34 & $+1.45$ & [$-0.52$, 3.35] \\
    saturation $0.50\times$ & 63.87 & $+0.98$ & [$-0.98$, 2.95] \\
    saturation $0.00\times$ & 63.82 & $+0.92$ & [$-0.98$, 2.83] \\
    \bottomrule
  \end{tabular}
\end{table}

\paragraph{Severe-level selection and confirmation.}
We selected a balanced panel of 180 MMMU-Pro rows, six from each of 30 subjects, by a fixed outcome-independent hash rule. The screen used noise $\sigma\in\{0.10,0.25,0.50,1.00\}$ and brightness factors
$\{1.5,3,6,12\}$. A level was carried forward when its paired loss was at least 15 points and its pointwise subject-stratified interval excluded zero. Noise $\sigma=1.0$ and brightness $12\times$ met this rule.

\begin{table}[htbp]
  \caption{Exploratory severe-corruption screen on the balanced 180-row
  MMMU-Pro panel. Accuracy and changes from the same-run clean arm count invalid
  generations as incorrect. Intervals are 10{,}000-replicate pointwise
  subject-stratified paired 95\% bootstrap intervals and are not adjusted for
  multiple comparisons. $\dagger$ marks the levels carried into disjoint
  confirmation.}
  \label{tab:glm53-severe-screen}
  \centering
  \small
  \begin{tabular}{lrrr}
    \toprule
    condition & accuracy (\%) & $\Delta$ vs.\ clean (pp) & paired 95\% CI \\
    \midrule
    clean & 61.11 & --- & --- \\
    noise $\sigma=0.10$ & 61.11 & $+0.00$ & [$-5.00$, 5.00] \\
    noise $\sigma=0.25$ & 55.56 & $-5.56$ & [$-10.56$, $-0.56$] \\
    noise $\sigma=0.50$ & 51.67 & $-9.44$ & [$-15.56$, $-3.33$] \\
    noise $\sigma=1.00$\,$^\dagger$ & 41.67 & $-19.44$ & [$-26.11$, $-12.78$] \\
    brightness $1.5\times$ & 60.00 & $-1.11$ & [$-5.56$, 3.33] \\
    brightness $3\times$ & 62.22 & $+1.11$ & [$-4.44$, 6.67] \\
    brightness $6\times$ & 49.44 & $-11.67$ & [$-17.78$, $-5.56$] \\
    brightness $12\times$\,$^\dagger$ & 44.44 & $-16.67$ & [$-23.33$, $-10.00$] \\
    \bottomrule
  \end{tabular}
\end{table}

We then generated corrupted responses for all 1,730 rows at these two levels. Because levels were selected using the 180-row screen, the primary confirmation excludes those rows and uses the remaining 1,550. The estimand is row-weighted paired accuracy with invalid generations counted as incorrect. Within-subject paired bootstrap resampling preserves the original subject weights. Two-sided 97.5\% Bonferroni intervals give familywise 95\% coverage across the two selected levels.

\begin{table}[htbp]
  \caption{Severe-corruption confirmation on the 1,550 MMMU-Pro rows disjoint
  from the severity-selection panel. Intervals are simultaneous familywise
  95\% intervals.}
  \label{tab:glm53-severe-confirmation}
  \centering
  \small
  \begin{tabular}{lrrr}
    \toprule
    condition & accuracy (\%) & $\Delta$ vs.\ clean (pp) & paired interval \\
    \midrule
    clean & 62.71 & --- & --- \\
    noise $\sigma=1.0$ & 41.29 & $-21.42$ & [$-24.52$, $-18.39$] \\
    brightness $12\times$ & 45.23 & $-17.48$ & [$-20.32$, $-14.58$] \\
    \bottomrule
  \end{tabular}
\end{table}

The moderate grid and severity screen are exploratory. The two severe levels are confirmed only for the stated MMMU-Pro population and corruption implementations.

\clearpage
\section{Attention-masking ablations: image-token mask and position}
\label{sec:ablations}

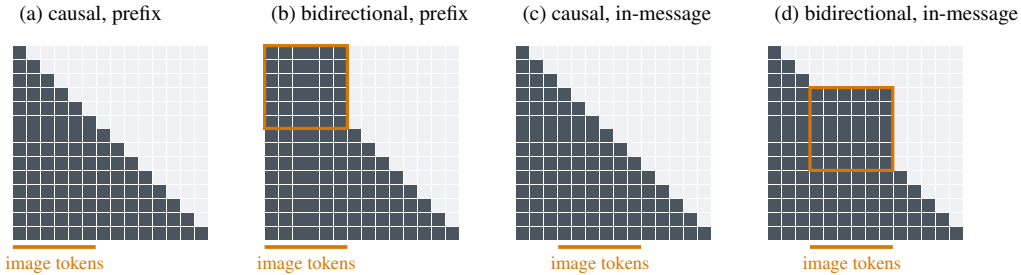
\begin{figure}[b]
  \centering
  \begin{tikzpicture}[font=\small, x=1.85mm, y=1.85mm]
    \definecolor{ablacc}{HTML}{D97706}
    \definecolor{ablon}{HTML}{4A5560}
    \definecolor{abloff}{HTML}{EFF1F4}
    \newcommand{\maskpanel}[4]{
    \begin{scope}[shift={(#1,0)}]
      \foreach \i in {0,...,13}{
        \foreach \j in {0,...,13}{
          \pgfmathsetmacro{\lo}{#2}
          \pgfmathsetmacro{\hi}{#2+6}
          \pgfmathparse{ (\j<=\i) ? 1 : ( #3==1 && \i>=\lo && \i<\hi && \j>=\lo && \j<\hi ? 1 : 0) }
          \ifdim\pgfmathresult pt>0.5pt
            \fill[ablon] (\j,-\i) rectangle ++(0.92,-0.92);
          \else
            \fill[abloff] (\j,-\i) rectangle ++(0.92,-0.92);
          \fi
        }
      }
      \ifnum#3=1
        \draw[ablacc, very thick] (#2,-#2) rectangle ++(5.92,-5.92);
      \fi
      \draw[ablacc, line width=1.4pt] (#2,-14.4) -- ++(5.92,0);
      \node[font=\scriptsize, text=ablacc] at (#2+3,-15.7) {image tokens};
      \node[font=\footnotesize, anchor=south west] at (-0.2,0.8) {#4};
    \end{scope}}
    \maskpanel{0}{0}{0}{(a) causal, prefix}
    \maskpanel{18}{0}{1}{(b) bidirectional, prefix}
    \maskpanel{36}{3}{0}{(c) causal, in-message}
    \maskpanel{54}{3}{1}{(d) bidirectional, in-message}
  \end{tikzpicture}
  \caption{\textbf{The mask$\times$placement grid.} Rows: query position;
  columns: key position; dark cells: attention permitted. Text tokens are
  causal in all variants; the orange square marks the mutually visible image
  block. (c) matches the reference convention.}
  \label{fig:abl-masks}
\end{figure}

These ablations change no architecture and no weights, that is, both axes only vary how attention is applied over the same tokens. By default the image tokens attend causally, like any other text token processed by an LLM, even though an image has no left-to-right order. The first axis therefore lets the image tokens attend to one another bidirectionally, at training and at inference. The second axis is whether the position of the image tokens matters: we can prepend them to the sequence (empirical evidence shows the first few tokens act as an attention sink~\citep{xiao2024efficient}), or replace the placeholder image tokens inside the user message, where chat templates place them, with the adapter latents. This gives a $2\times2$ grid of \{mask $\times$ position\} (Figure~\ref{fig:abl-masks}), trained at the
Qwen 14B reader on the reference recipe, one run per cell; 14B is the largest scale at which a full grid was within compute reach.

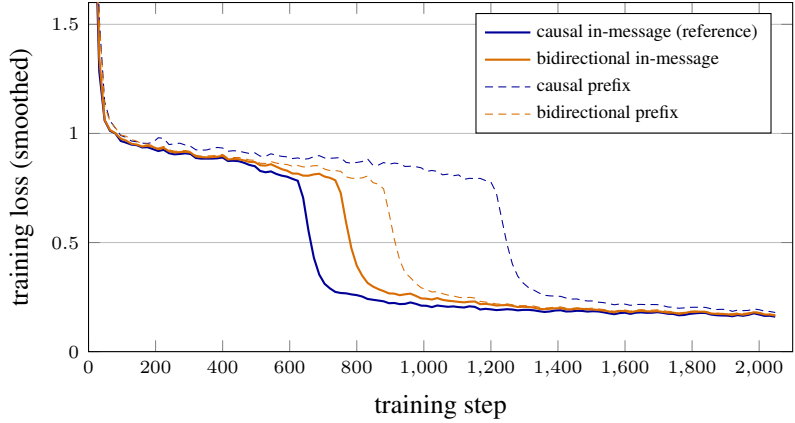
\begin{figure}[!t]
  \centering
  \begin{tikzpicture}
    \begin{axis}[
        width=0.78\textwidth, height=6.2cm,
        xmin=0, xmax=2100, ymin=0, ymax=1.6,
        xlabel={training step}, ylabel={training loss (smoothed)},
        ymajorgrids,
        tick label style={font=\scriptsize},
        legend style={font=\scriptsize, at={(0.97,0.97)}, anchor=north east},
        legend cell align=left]
      \addplot[blue!60!black, thick] coordinates { (16,2.1851) (32,1.2846) (48,1.06) (64,1.0121) (80,1.0001) (96,0.9656) (112,0.9578) (128,0.9488) (144,0.9461) (160,0.936) (176,0.937) (192,0.9283) (208,0.9194) (224,0.9249) (240,0.9095) (256,0.9042) (272,0.9059) (288,0.9089) (304,0.906) (320,0.8875) (336,0.8846) (352,0.8847) (368,0.8848) (384,0.8861) (400,0.8898) (416,0.8736) (432,0.8749) (448,0.8728) (464,0.8654) (480,0.8555) (496,0.8498) (512,0.828) (528,0.822) (544,0.826) (560,0.8143) (576,0.8066) (592,0.8019) (608,0.792) (624,0.7831) (640,0.7055) (656,0.5536) (672,0.4305) (688,0.3534) (704,0.3109) (720,0.2899) (736,0.2737) (752,0.2687) (768,0.2672) (784,0.2623) (800,0.259) (816,0.2526) (832,0.2415) (848,0.2378) (864,0.2343) (880,0.2307) (896,0.222) (912,0.223) (928,0.2178) (944,0.2189) (960,0.2254) (976,0.2192) (992,0.2109) (1008,0.2101) (1024,0.2041) (1040,0.2114) (1056,0.2083) (1072,0.2055) (1088,0.2073) (1104,0.204) (1120,0.2023) (1136,0.2077) (1152,0.2052) (1168,0.1932) (1184,0.1961) (1200,0.1938) (1216,0.1904) (1232,0.1921) (1248,0.1944) (1264,0.193) (1280,0.1892) (1296,0.1913) (1312,0.1915) (1328,0.1893) (1344,0.1852) (1360,0.1818) (1376,0.1841) (1392,0.1886) (1408,0.1896) (1424,0.184) (1440,0.1846) (1456,0.1872) (1472,0.1866) (1488,0.1836) (1504,0.1852) (1520,0.1842) (1536,0.1803) (1552,0.1756) (1568,0.1716) (1584,0.1767) (1600,0.1783) (1616,0.1731) (1632,0.1784) (1648,0.1799) (1664,0.1778) (1680,0.1782) (1696,0.1808) (1712,0.1779) (1728,0.1735) (1744,0.1726) (1760,0.1667) (1776,0.168) (1792,0.1729) (1808,0.1755) (1824,0.1774) (1840,0.1772) (1856,0.1701) (1872,0.167) (1888,0.1657) (1904,0.1651) (1920,0.1623) (1936,0.1674) (1952,0.1643) (1968,0.1684) (1984,0.1757) (2000,0.1702) (2016,0.1639) (2032,0.1644) (2048,0.1602) };
      \addplot[orange!85!black, thick] coordinates { (16,2.3669) (32,1.3502) (48,1.0623) (64,1.0156) (80,1.0013) (96,0.9762) (112,0.9672) (128,0.9526) (144,0.9488) (160,0.9404) (176,0.9436) (192,0.9382) (208,0.9277) (224,0.9337) (240,0.9184) (256,0.9133) (272,0.9142) (288,0.9161) (304,0.9126) (320,0.8946) (336,0.8916) (352,0.8903) (368,0.8914) (384,0.8933) (400,0.8975) (416,0.8806) (432,0.8828) (448,0.8818) (464,0.8762) (480,0.8695) (496,0.869) (512,0.8541) (528,0.8527) (544,0.8589) (560,0.8476) (576,0.8363) (592,0.8294) (608,0.8164) (624,0.8158) (640,0.8067) (656,0.8063) (672,0.8124) (688,0.8152) (704,0.8024) (720,0.7965) (736,0.7854) (752,0.7277) (768,0.5927) (784,0.4735) (800,0.3939) (816,0.3487) (832,0.3151) (848,0.2986) (864,0.2867) (880,0.2773) (896,0.267) (912,0.2658) (928,0.2586) (944,0.2628) (960,0.2658) (976,0.2556) (992,0.244) (1008,0.2426) (1024,0.2389) (1040,0.2453) (1056,0.2374) (1072,0.2334) (1088,0.2309) (1104,0.228) (1120,0.2259) (1136,0.2288) (1152,0.2299) (1168,0.2177) (1184,0.2189) (1200,0.2161) (1216,0.2113) (1232,0.2128) (1248,0.2137) (1264,0.2098) (1280,0.2046) (1296,0.2055) (1312,0.2046) (1328,0.2018) (1344,0.1962) (1360,0.1945) (1376,0.1966) (1392,0.1985) (1408,0.1993) (1424,0.1942) (1440,0.1938) (1456,0.1966) (1472,0.1959) (1488,0.1934) (1504,0.1934) (1520,0.1908) (1536,0.1883) (1552,0.1841) (1568,0.1799) (1584,0.1866) (1600,0.186) (1616,0.1805) (1632,0.1853) (1648,0.1868) (1664,0.1849) (1680,0.184) (1696,0.1855) (1712,0.1821) (1728,0.1788) (1744,0.1787) (1760,0.1733) (1776,0.1765) (1792,0.1813) (1808,0.1819) (1824,0.1833) (1840,0.182) (1856,0.1748) (1872,0.1731) (1888,0.1721) (1904,0.1714) (1920,0.1678) (1936,0.1728) (1952,0.1692) (1968,0.1731) (1984,0.1806) (2000,0.1764) (2016,0.1703) (2032,0.1695) (2048,0.1656) };
      \addplot[blue!60!black, densely dashed] coordinates { (16,2.506) (32,1.5262) (48,1.1456) (64,1.0582) (80,1.0249) (96,0.9907) (112,0.982) (128,0.9653) (144,0.9618) (160,0.9554) (176,0.9543) (192,0.9588) (208,0.9802) (224,0.9775) (240,0.9492) (256,0.9522) (272,0.9554) (288,0.9472) (304,0.9415) (320,0.925) (336,0.9234) (352,0.9242) (368,0.9311) (384,0.9358) (400,0.9332) (416,0.9184) (432,0.9208) (448,0.9191) (464,0.9132) (480,0.9081) (496,0.9111) (512,0.8958) (528,0.8952) (544,0.9068) (560,0.8996) (576,0.8951) (592,0.8926) (608,0.8849) (624,0.8878) (640,0.8815) (656,0.8842) (672,0.8937) (688,0.8983) (704,0.8877) (720,0.8861) (736,0.8876) (752,0.8894) (768,0.8672) (784,0.8637) (800,0.8671) (816,0.8712) (832,0.8836) (848,0.8544) (864,0.8548) (880,0.8667) (896,0.8586) (912,0.8641) (928,0.8613) (944,0.8486) (960,0.8497) (976,0.8476) (992,0.8425) (1008,0.8261) (1024,0.8303) (1040,0.826) (1056,0.8104) (1072,0.8135) (1088,0.8063) (1104,0.8099) (1120,0.7898) (1136,0.7943) (1152,0.7931) (1168,0.794) (1184,0.784) (1200,0.7772) (1216,0.7286) (1232,0.6107) (1248,0.4931) (1264,0.4019) (1280,0.346) (1296,0.3149) (1312,0.2936) (1328,0.2821) (1344,0.2686) (1360,0.2583) (1376,0.2556) (1392,0.2545) (1408,0.2533) (1424,0.2458) (1440,0.2408) (1456,0.2415) (1472,0.2355) (1488,0.2322) (1504,0.2322) (1520,0.227) (1536,0.2245) (1552,0.2202) (1568,0.2147) (1584,0.2184) (1600,0.2178) (1616,0.2122) (1632,0.2158) (1648,0.2165) (1664,0.2151) (1680,0.2159) (1696,0.2155) (1712,0.2111) (1728,0.2046) (1744,0.2034) (1760,0.1988) (1776,0.2004) (1792,0.203) (1808,0.2034) (1824,0.2036) (1840,0.2027) (1856,0.1968) (1872,0.193) (1888,0.194) (1904,0.1931) (1920,0.1846) (1936,0.1903) (1952,0.1878) (1968,0.1891) (1984,0.1957) (2000,0.1922) (2016,0.1856) (2032,0.1838) (2048,0.1794) };
      \addplot[orange!85!black, densely dashed] coordinates { (16,2.6173) (32,1.5203) (48,1.1255) (64,1.0562) (80,1.026) (96,0.9932) (112,0.9871) (128,0.9722) (144,0.9648) (160,0.9515) (176,0.9515) (192,0.9433) (208,0.9328) (224,0.9402) (240,0.9252) (256,0.9187) (272,0.9193) (288,0.9218) (304,0.9179) (320,0.8993) (336,0.8958) (352,0.8969) (368,0.8977) (384,0.8999) (400,0.9037) (416,0.8868) (432,0.8894) (448,0.8898) (464,0.8846) (480,0.8783) (496,0.877) (512,0.8632) (528,0.8626) (544,0.8711) (560,0.864) (576,0.8598) (592,0.8575) (608,0.8496) (624,0.8531) (640,0.8466) (656,0.848) (672,0.8522) (688,0.8521) (704,0.8366) (720,0.8298) (736,0.8297) (752,0.8264) (768,0.8014) (784,0.7948) (800,0.794) (816,0.7963) (832,0.8048) (848,0.7747) (864,0.7676) (880,0.746) (896,0.6368) (912,0.5151) (928,0.4197) (944,0.3616) (960,0.3393) (976,0.3125) (992,0.2928) (1008,0.2842) (1024,0.2734) (1040,0.2749) (1056,0.2657) (1072,0.259) (1088,0.2552) (1104,0.2491) (1120,0.2465) (1136,0.2489) (1152,0.2457) (1168,0.2304) (1184,0.2271) (1200,0.2233) (1216,0.2185) (1232,0.2172) (1248,0.2189) (1264,0.2148) (1280,0.2107) (1296,0.2107) (1312,0.2101) (1328,0.2089) (1344,0.2031) (1360,0.1996) (1376,0.201) (1392,0.205) (1408,0.2049) (1424,0.1993) (1440,0.1979) (1456,0.2007) (1472,0.2001) (1488,0.1976) (1504,0.1995) (1520,0.1961) (1536,0.1929) (1552,0.1892) (1568,0.1848) (1584,0.1903) (1600,0.1909) (1616,0.1856) (1632,0.1907) (1648,0.1924) (1664,0.1901) (1680,0.1899) (1696,0.1917) (1712,0.188) (1728,0.1845) (1744,0.184) (1760,0.177) (1776,0.1799) (1792,0.1856) (1808,0.1863) (1824,0.1858) (1840,0.1843) (1856,0.1774) (1872,0.1768) (1888,0.1753) (1904,0.1753) (1920,0.1735) (1936,0.1775) (1952,0.1749) (1968,0.1776) (1984,0.1841) (2000,0.1797) (2016,0.1732) (2032,0.1725) (2048,0.169) };
      \legend{causal in-message (reference), bidirectional in-message, causal prefix, bidirectional prefix}
    \end{axis}
  \end{tikzpicture}
  \caption{\textbf{A shared plateau, entered by every cell; the cells differ in when they transition.} Training loss (window-25 smoothed) for the $2\times2$ grid at the 14B reader, one run per cell. All four runs sit on a plateau between roughly $0.95$ and $0.80$; transitions are abrupt ($\sim$40--50 steps); the reference convention transitions first.}
  \label{fig:abl-metastable}
\end{figure}

All four variations' loss curves follow the same two-phase dynamic (Figure~\ref{fig:abl-metastable}): the loss drops quickly to a plateau, drifts slowly from about $0.95$ down to $0.80$, and then drops sharply to the converged regime near $0.16$--$0.18$ --- a grokking-like delayed transition~\citep{power2022grokking}, here in the training loss itself. The transition is abrupt: once a run leaves the plateau it reaches loss $0.5$ within roughly $40$--$50$ steps. What differs between cells is how long they remain on the plateau. The reference cell (causal, in-message) transitioned first, at step $647$; bidirectional in-message at $765$; bidirectional prefix at $895$; and causal prefix at $1230$, the last after sitting on the plateau for over a thousand steps.

We compared plateau and post-transition checkpoints of the same runs through the frozen reader on 16 fixed examples. The plateau is not the model ignoring the image: with the image removed, loss is $4.01$ versus $0.84$--$0.91$ on the plateau, so the transition sharpens image information that is already being used. Nor is it the attention sink~\citep{xiao2024efficient}: sink geometry is fixed by the placement choice, yet all cells plateau at a similar level. Our reading is that, with attention frozen, the plateau is the MLP projector not yet packing the latents into a form interpretable to the frozen reader, which is consistent with the final checkpoints' evaluations differing little across cells (Table~\ref{tab:abl-evalgrid}).

\FloatBarrier

Evaluations use the generative protocol of Table~\ref{tab:rung-rl} without RL (greedy decoding, the official answer parser, seeded random fallback on parse failure, counted and reported), with reasoning mode off. Table~\ref{tab:abl-evalgrid} reports all four cells at their final SFT
checkpoints.

\begin{table}[htbp]
  \caption{Eval grid for the $2\times2$ ablation, final SFT checkpoints
  (no RL), reasoning mode off, generative protocol, accuracy \%
  (parse-fallback \% in parentheses).}
  \label{tab:abl-evalgrid}
  \centering
  \scriptsize
  \begin{tabular}{lccc}
    \toprule
    cell (14B reader) & MMMU & MMMU-Pro & BLINK6 \\
    \midrule
    causal, in-message (reference) & 55.3 (0.0) & 33.2 (0.0) & 49.7 (0.0) \\
    bidirectional, in-message & 53.4 (0.0) & 33.6 (0.0) & 48.6 (0.0) \\
    bidirectional, prefix & 51.8 (0.0) & 32.8 (0.2) & 48.8 (0.0) \\
    causal, prefix & 54.0 (0.8) & 32.5 (0.2) & 48.3 (0.0) \\
    \bottomrule
  \end{tabular}
\end{table}

With one run per cell these orderings are observations, not laws. What the grid supports is that the configuration mostly decided when the transition happened, not where training ends up; the reference convention is a safe default.
\clearpage
\section{Interpretability details}
\label{app:interp}

\subsection{Per-patch word reading}
\label{app:interp-winner}
Both lenses in Figure~\ref{fig:interp-grass} read one image-token position at a time. The Jacobian-lens direction is $v_{w,\ell} = \mathbb{E}\,[\partial (u_w^\top h_{\mathrm{final}}) / \partial h_\ell]$, the word's unembedding row pulled back through the remaining layers, averaged over 23 images passed through the adapter.\footnote{Computed through GLM by adding input-gradient formulas to its quantized inference kernels.} At layer $\ell$ a candidate word $w$ scores $u_w^\top h_\ell$ under the logit lens and
$v_{w,\ell}^\top h_\ell$ under the Jacobian lens; the higher-scoring candidate wins the patch. Candidate words are single-token forms with a leading space (`` dog'', `` grass'', `` cat''): ``dog'' against ``grass'' for the maps, ``dog'' against ``cat'' for the control below.

Patches are $28\times28$-pixel cells (ViT patch 14 with a $2\times2$ merge; images resized so the long side is at most 448, e.g.\ a $14\times16$ grid in Figure~\ref{fig:interp-grass}).

Dog masks on the two grass-background images are hand-annotated; the control images use COCO instance annotations converted to token cells. A cell counts as grass when its mean $G{-}R$ exceeds 8 and mean $G{-}B$ exceeds 15 on the 0--255 scale, excluding cells on the dog mask.

The Figure~\ref{fig:interp-grass} curves average the two grass-background images; the control takes the median over six dog and six cat images.

\subsection{Bias-controlled reading test}
\label{app:interp-control}
The dog-vs-grass maps in Figure~\ref{fig:interp-grass} label grass patches by a color heuristic, and two-word winner maps can be moved by the reader's word priors alone. The control below removes both concerns: on real COCO masks, we score how often ``dog'' beats ``cat'' on dog patches of six dog images minus how often it falsely beats ``cat'' on cat patches of six cat images (a dog-less prior control); we call this difference the discrimination margin. The Jacobian lens discriminates from the first layer; the logit lens is weak and unstable before layer $\approx$50, dipping to zero near layers 25 and 41--43. The word-prior effect is visible directly in this control: on the dog-less images, ``dog'' falsely beats ``cat'' on the
cat's own patches until layer $\approx$45 under the logit lens. On the post-RL Qwen stitches the same test finds no comparable early channel (not shown): Jacobian-lens discrimination is weak and decays with depth at 14B and 8B, appears only in the last layers at 4B, and the logit lens is the stronger late channel for all three.

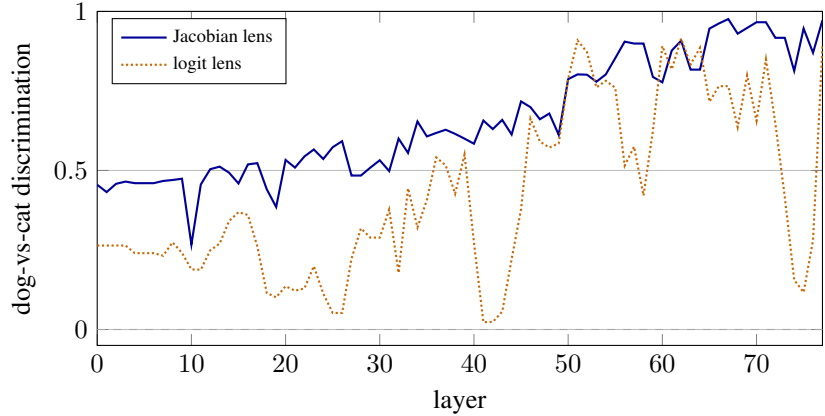
\begin{figure}[h]
  \centering
  \begin{tikzpicture}
    \begin{axis}[
        width=0.8\textwidth, height=6cm,
        xmin=0, xmax=77,
        xlabel={layer},
        ylabel={dog-vs-cat discrimination},
        ymin=-0.05, ymax=1, ymajorgrids,
        legend style={at={(0.02,0.98)}, anchor=north west, font=\scriptsize},
        legend cell align=left]
      \addplot[blue!60!black, thick, mark=none] coordinates {
        (0,0.455) (1,0.432) (2,0.458) (3,0.465) (4,0.460) (5,0.460) (6,0.460) (7,0.467) (8,0.470) (9,0.474) (10,0.267) (11,0.456) (12,0.504) (13,0.512) (14,0.493) (15,0.459) (16,0.519) (17,0.523) (18,0.441) (19,0.385) (20,0.533) (21,0.509) (22,0.544) (23,0.566) (24,0.536) (25,0.573) (26,0.592) (27,0.484) (28,0.484) (29,0.509) (30,0.532) (31,0.498) (32,0.600) (33,0.555) (34,0.654) (35,0.607) (36,0.618) (37,0.628) (38,0.615) (39,0.600) (40,0.584) (41,0.657) (42,0.630) (43,0.659) (44,0.613) (45,0.717) (46,0.699) (47,0.661) (48,0.679) (49,0.612) (50,0.787) (51,0.802) (52,0.801) (53,0.779) (54,0.802) (55,0.853) (56,0.905) (57,0.899) (58,0.899) (59,0.794) (60,0.777) (61,0.877) (62,0.908) (63,0.817) (64,0.817) (65,0.946) (66,0.962) (67,0.976) (68,0.930) (69,0.948) (70,0.966) (71,0.966) (72,0.917) (73,0.917) (74,0.813) (75,0.946) (76,0.870) (77,0.972)};
      \addplot[orange!80!black, thick, densely dotted, mark=none] coordinates {
        (0,0.264) (1,0.264) (2,0.264) (3,0.264) (4,0.240) (5,0.240) (6,0.240) (7,0.232) (8,0.274) (9,0.241) (10,0.189) (11,0.188) (12,0.249) (13,0.270) (14,0.341) (15,0.369) (16,0.360) (17,0.261) (18,0.115) (19,0.101) (20,0.136) (21,0.122) (22,0.129) (23,0.199) (24,0.113) (25,0.052) (26,0.052) (27,0.223) (28,0.318) (29,0.289) (30,0.289) (31,0.377) (32,0.178) (33,0.444) (34,0.321) (35,0.407) (36,0.541) (37,0.516) (38,0.427) (39,0.550) (40,0.273) (41,0.023) (42,0.024) (43,0.056) (44,0.221) (45,0.378) (46,0.663) (47,0.591) (48,0.572) (49,0.585) (50,0.788) (51,0.909) (52,0.872) (53,0.761) (54,0.782) (55,0.760) (56,0.515) (57,0.575) (58,0.421) (59,0.625) (60,0.891) (61,0.818) (62,0.915) (63,0.833) (64,0.886) (65,0.716) (66,0.765) (67,0.765) (68,0.636) (69,0.799) (70,0.656) (71,0.849) (72,0.648) (73,0.417) (74,0.156) (75,0.117) (76,0.281) (77,0.891)};
      \addplot[dashed, gray!70, domain=0:77] {0};
      \legend{Jacobian lens, logit lens}
    \end{axis}
  \end{tikzpicture}
  \caption{\textbf{Bias-controlled reading: the adapter's code is readable in gradient coordinates from the first layers; in embedding coordinates only late.} Dog-vs-cat discrimination margin on masked patches (released 743B stitch), twelve COCO images, median. The Jacobian lens (solid) discriminates at every depth; the logit lens (dotted) tracks the word prior and is unreliable before $\approx$ layer 50.}
  \label{fig:interp-discrim}
\end{figure}

\subsection{Steering methodology}
\label{app:steering}
All interventions add a unit-norm direction to the residual stream at the image-token positions during the prefill, at layer $\ell$ and dose $\alpha$ times the mean residual norm of the perturbed positions (so equal $\alpha$ means an equal relative perturbation at every layer). Answers are $k = 12$--$16$-token greedy generations. A cell is one (image, layer) pair. The 24 cells are 12 images steered at two layers each. 

Formally, with $P$ the perturbed positions (the object-mask image tokens unless an arm says otherwise) and $h_{\ell,t}$ the residual stream entering layer $\ell$ at position $t$:

\begin{quote}
\textbf{Algorithm} (residual-stream steering at layer $\ell$, dose $\alpha$).
\emph{Input:} the spliced sequence (image tokens $Y$ at positions
$P_{\mathrm{img}}$, prompt tokens), direction $v$, layer $\ell$, dose
$\alpha$, positions $P$.
\begin{enumerate}
  \item $\hat v \leftarrow v / \lVert v \rVert_2$.
  \item Prefill the frozen reader, intercepting the residual stream entering
    layer $\ell$: set the layer-calibrated scale
    $s \leftarrow \alpha \cdot \frac{1}{|P|} \sum_{t \in P}
    \lVert h_{\ell,t} \rVert_2$, then
    $h_{\ell,t} \leftarrow h_{\ell,t} + s\,\hat v$ for every $t \in P$.
    Decode steps are untouched; the edit persists through the key--value
    cache.
  \item Record the logits at the last prompt position; greedy-decode $k$
    tokens.
\end{enumerate}
\end{quote}

\textbf{Arms.} The six arms test whether the effect is specific to the Jacobian direction and to the image positions. All directions are unit-normalized before step~1, and all arms share the layer, dose, and decode. The intervention arm, \emph{inject}, adds the Jacobian-lens direction of the target word, $v_{w_{\mathrm{tgt}},\ell}$ (Section~\ref{sec:interp}, averaged over 23 images), at the object-mask
image tokens. Three control arms hold the positions fixed and vary the direction: \emph{random} adds a Gaussian direction $v \sim \mathcal{N}(0, I)$, which controls for dose alone; \emph{unembedding} adds the target word's unembedding row $u_{w_{\mathrm{tgt}}}$, the direction the logit lens reads with, which tests whether the output basis suffices for writing; and \emph{class-mean} adds $\mu_{\mathrm{tgt}}(\ell) - \mu_{\mathrm{src}}(\ell)$, where $\mu_c(\ell)$ is the mean image-token activation at layer $\ell$ over held-out images of class $c$, the standard activation-steering baseline. Two control arms hold the direction fixed and vary the positions: \emph{background} adds the inject direction at background patches rather than the object, testing spatial specificity, and \emph{text-position} adds it at the question's text tokens rather than the image tokens, testing whether the direction acts on the percept or on the text stream.

\textbf{Criteria.} A \emph{generation flip} requires the greedy answer to gain the target word and drop the source word; a \emph{logit crossover} requires the target word's logit to exceed the source word's at the answer position. Images whose unperturbed answer does not name the source class are excluded from generation-flip rates. Where baseline margins are small the crossover criterion admits noise crossings (random directions cross on the easiest images), so generation flips are the conservative measure. Under injection at $\alpha{=}0.2$ the model's color reports stay correct in 11 of 12 cells; one answer flips the class while keeping the attribute (``Brown \ldots\ the cat is''). Per-image and per-position variants of the Jacobian directions strengthen steering where it works and do not rescue it where it fails. Reading and writing can dissociate: on the Qwen stitches the same directions can read while no dose changes the output, so steering is convergent rather than conclusive evidence for the lens.

\subsubsection{The same directions steer GLM; they do not steer the smaller
stitches.}
The Jacobian directions are causal levers as well as probes. That is, adding them to the residual stream changes the model's answer. On a dog image, we add $v_{\mathrm{cat},\ell}$ at the image-token positions of layer $\ell$, scaled to a dose $\alpha$ times the local residual norm. At $\alpha=0.2$ the ``cat'' logit overtakes ``dog'' in 22--24 of 24 cells, and the generated answers change to ``cat'' coherently. This replicates in the reverse direction (cat to dog), with a second question phrasing, with other object categories, and under GLM's own chat template. Three control directions (random, the unembedding row, and the class-mean difference) flip at most one cell in 24 at the same dose. Adding the direction at the question's text positions instead of the image positions flips none, so the edit acts on the percept rather than on the text. The steered model still reports the dog's color correctly in 11 of 12 cells, so the intervention supposedly changes the object category without destroying the rest of the percept. The identical battery on the post-RL Qwen3 systems (4B, 8B, 14B) flips at most 1 of 12 cells at any dose, on images they answer correctly under their own templates, and their Jacobian-lens reading is similarly weak. Behavioral evaluation does not distinguish these cases: all four models answer correctly, and only intervention separates them. In this series, early readability and steerability appear together only at the frontier model. Note that we don't consider this to be a claim about scaling, because the Qwen3 models differ from GLM in reader family and reinforcement task as well as in size, and four models cannot separate these explanations.

\subsection{Text--image token geometry}
\label{app:interp-geometry}
\begin{figure}[h]
  \centering
  \begin{tikzpicture}
    \begin{axis}[
        width=0.72\textwidth, height=5.2cm,
        xmin=-1, xmax=77,
        xlabel={layer}, ylabel={text--image centroid cosine},
        ymin=0, ymax=1, ymajorgrids,
        legend style={at={(0.98,0.05)}, anchor=south east, font=\scriptsize}]
      \addplot[blue!60!black, thick, mark=none] coordinates {
        (-1,0.017) (0,0.117) (1,0.148) (2,0.173) (3,0.178) (4,0.155) (5,0.153) (6,0.155) (7,0.185) (8,0.240) (9,0.282) (10,0.397) (11,0.460) (12,0.460) (13,0.492) (14,0.527) (15,0.561) (16,0.534) (17,0.575) (18,0.594) (19,0.554) (20,0.586) (21,0.570) (22,0.597) (23,0.598) (24,0.634) (25,0.642) (26,0.641) (27,0.628) (28,0.643) (29,0.637) (30,0.638) (31,0.651) (32,0.653) (33,0.655) (34,0.652) (35,0.652) (36,0.640) (37,0.642) (38,0.661) (39,0.690) (40,0.700) (41,0.700) (42,0.703) (43,0.710) (44,0.709) (45,0.716) (46,0.738) (47,0.724) (48,0.734) (49,0.750) (50,0.756) (51,0.762) (52,0.760) (53,0.769) (54,0.758) (55,0.785) (56,0.780) (57,0.777) (58,0.760) (59,0.760) (60,0.762) (61,0.771) (62,0.768) (63,0.787) (64,0.809) (65,0.794) (66,0.793) (67,0.795) (68,0.801) (69,0.815) (70,0.815) (71,0.838) (72,0.832) (73,0.831) (74,0.836) (75,0.851) (76,0.860) (77,0.911)};
      \addplot[dashed, gray!70] coordinates {(12,0) (12,1)};
      \legend{text--image centroid cosine, layer 12}
    \end{axis}
  \end{tikzpicture}\\[4pt]
  \includegraphics[width=0.98\textwidth]{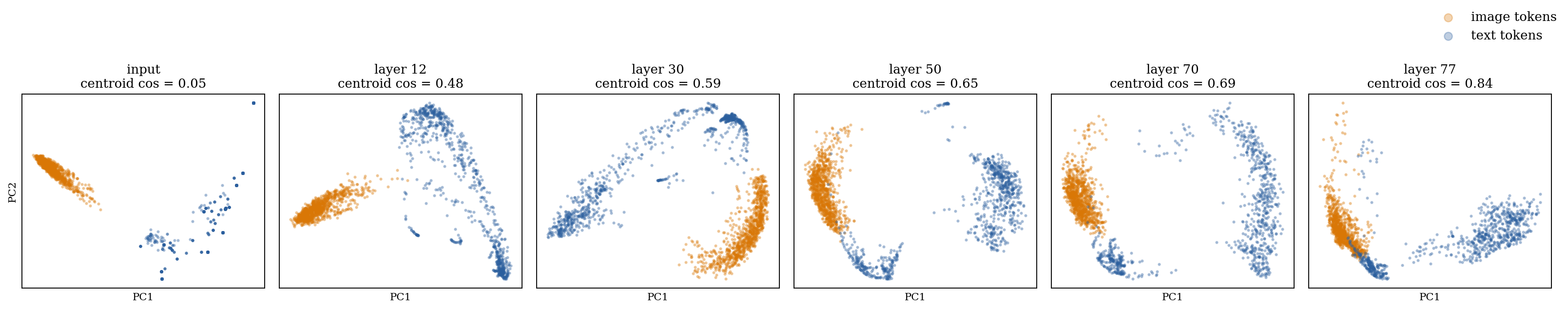}
  \caption{\textbf{Image and text tokens enter the frozen reader nearly
  orthogonal and merge gradually with depth.} Top: cosine similarity between
  the mean image token and the mean text token per layer (unit-normalized
  tokens; 18 COCO images, released 743B stitch; $x{=}-1$ is the spliced
  input); half of the total rise is reached by layer 12. Bottom: joint
  per-layer PCA of image-token and text-token hidden states (31 images,
  unit-normalized residuals; same model and splice, but a separate image set
  from the curve, so the per-panel centroid cosines differ slightly from the
  curve at the same layers). The populations are disjoint at the input and
  begin to overlap by layer 77 while remaining largely separable.}
  \label{fig:interp-gap-pca}
\end{figure}
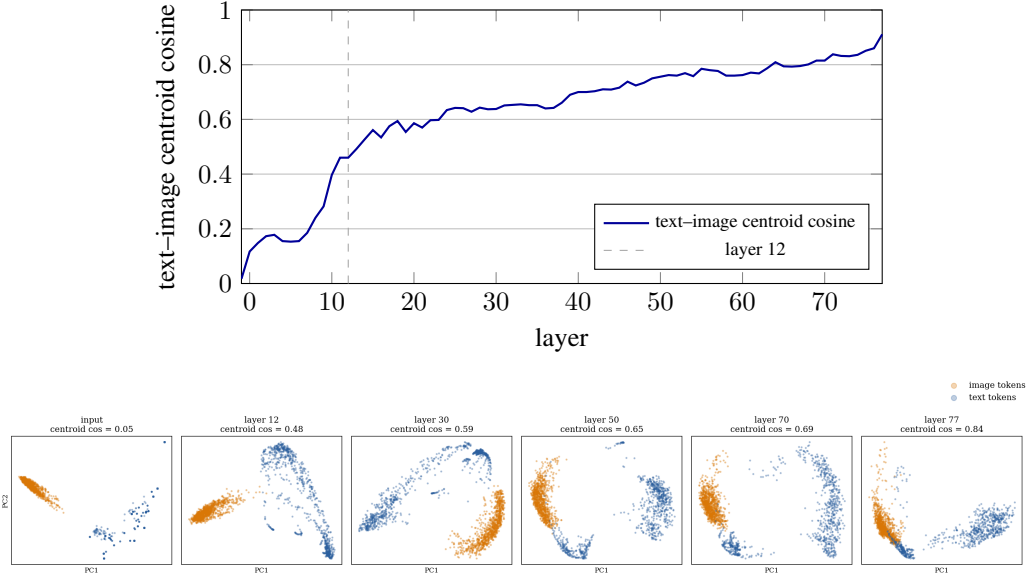

We measure how far the adapter's image tokens are from the reader's text tokens, and how that distance changes with depth. At each layer we take the mean hidden state of the image tokens and the mean hidden state of the text tokens and compute the cosine between them (Figure~\ref{fig:interp-gap-pca}, top). At the input the two means are nearly orthogonal (cosine 0.02). The cosine rises to 0.46 by layer 12 and to 0.91 at the final layer 77. A per-layer PCA of the two token populations (Figure~\ref{fig:interp-gap-pca}, bottom) shows the same progression: the image and text tokens start as two separate clusters, move toward each other with depth, and partly overlap by the last layer without merging.

The adapter's output tokens also do not resemble any individual word embedding. For each image token at the input we compute its maximum cosine to any row of the reader's vocabulary embedding matrix, after subtracting the mean embedding from both. Averaged over image tokens, this maximum is 0.059. Random vectors with the same norms give 0.057. The adapter therefore does not write its output in the reader's word coordinates, which is why the logit lens reads it poorly at early layers (Appendix~\ref{app:interp-control}).

\clearpage

\section{Compute details}
\label{app:compute}

All experiments account for approximately 4,800 NVIDIA H100 GPU-hours, 940 H200 GPU-hours, and 120 B200 GPU-hours. These estimates include resumed training segments. Additional preliminary, failed, and diagnostic work required further compute not included in these totals.


\newpage

\end{document}